\documentclass[journal]{IEEEtran}
\ifCLASSINFOpdf
\else
\fi

\usepackage{algorithm}
\usepackage{algorithmic}
\usepackage{multirow}
\usepackage{latexsym}
\usepackage{subfigure}
\usepackage{graphicx}
\usepackage{amsmath}
\usepackage{amssymb}
\usepackage{subfigure}

\usepackage{hyperref}
\hypersetup{hypertex=true,
	colorlinks=true,
	linkcolor=red,
	anchorcolor=red,
	citecolor=green}

\begin{document}
%
\title{SGMNet: Scene Graph Matching Network for Few-Shot Remote Sensing Scene Classification}
%
%
%

\author{Baoquan~Zhang, 
	    Shanshan~Feng, 
	    Xutao~Li, 
	    Yunming~Ye, 
	    Rui~Ye 
\thanks{Baoquan Zhang, Shanshan Feng, Xutao Li, Yunming Ye, and Rui Ye are with the School of Computer Science and Technology, Harbin Institute of Technology, Shenzhen, Shenzhen 518055, Guangdong, China.}
\thanks{E-mail: zhangbaoquan@stu.hit.edu.cn, victor\_fengss@foxmail.com, \{lixutao, yeyunming\}@hit.edu.cn, yerui\_hitsz@163.com}
\thanks{Corresponding author: Yunming Ye.\protect\\}
\thanks{Manuscript received September 15, 2021.}}

%
%

\markboth{Journal of \LaTeX\ Class Files,~Vol.~13, No.~9, September~2014}%
{Shell \MakeLowercase{\textit{et al.}}: Bare Demo of IEEEtran.cls for Journals}
%



\maketitle

\begin{abstract}
Few-Shot Remote Sensing Scene Classification (FSRSSC) is an important task, which aims to recognize novel scene classes with few examples. Recently, several studies attempt to address the FSRSSC problem by following few-shot natural image classification methods. These existing methods have made promising progress and achieved superior performance. However, they all overlook two unique characteristics of remote sensing images: (\romannumeral1) \emph{object co-occurrence} that multiple objects tend to appear together in a scene image and (\romannumeral2) \emph{object spatial correlation} that these co-occurrence objects are distributed in the scene image following some spatial structure patterns. Such unique characteristics are very beneficial for FSRSSC, which can effectively alleviate the scarcity issue of labeled remote sensing images since they can provide more refined descriptions for each scene class. To fully exploit these characteristics, we propose a novel scene graph matching-based meta-learning framework for FSRSSC, called SGMNet. In this framework, a scene graph construction module is carefully designed to represent each test remote sensing image or each scene class as a scene graph, where the nodes reflect these co-occurrence objects meanwhile the edges capture the spatial correlations between these co-occurrence objects. Then, a scene graph matching module is further developed to evaluate the similarity score between each test remote sensing image and each scene class. Finally, based on the similarity scores, we perform the scene class prediction via a nearest neighbor classifier. We conduct extensive experiments on UCMerced LandUse, WHU19, AID, and NWPU-RESISC45 datasets. The experimental results show that our method obtains superior performance over the previous state-of-the-art methods.
\end{abstract}


%
\IEEEpeerreviewmaketitle

\section{Introduction}
\label{section1}
%
%
%
%
\IEEEPARstart{R}{emote} sensing scene classification (RSSC) is an important research problem on the remote sensing applications, such as disaster detection \cite{amit2017disaster}, urban planning \cite{pham2011case}, land use \cite{joshi2016review}, etc. It aims to recognize a remote sensing image to a scene semantic class. During the past years, RSSC has made rapid advances such as \cite{xie2019scale, sun2019remote}, relying on convolutional neural network (CNN) and a large number of labeled data.
However, unlike the annotation task of natural images \cite{HeZRS16}, the annotation task of remote sensing images is very time-consuming and laborious since it needs to be achieved by experts with rich knowledge of remote sensing domains \cite{2020RSmetanet}. 
To alleviate the burden of annotating data, the problem of \underline{F}ew-\underline{S}hot \underline{R}emote \underline{S}ensing \underline{S}cene \underline{C}lassification (FSRSSC) has been proposed recently, which draws wide attention. Different from RSSC, FSRSSC aims to learn task-agnostic meta-knowledge from base scene classes with abundant labeled samples and then transfer the meta-knowledge to recognize novel scene classes with limited labeled samples \cite{2020RSmetanet, 2020FewAlajaji, cheng2021spnet}, where these novel scene classes are not overlapped with the base scene classes.

\begin{figure}[!t]
	\centering
	\includegraphics[width=1.00\columnwidth]{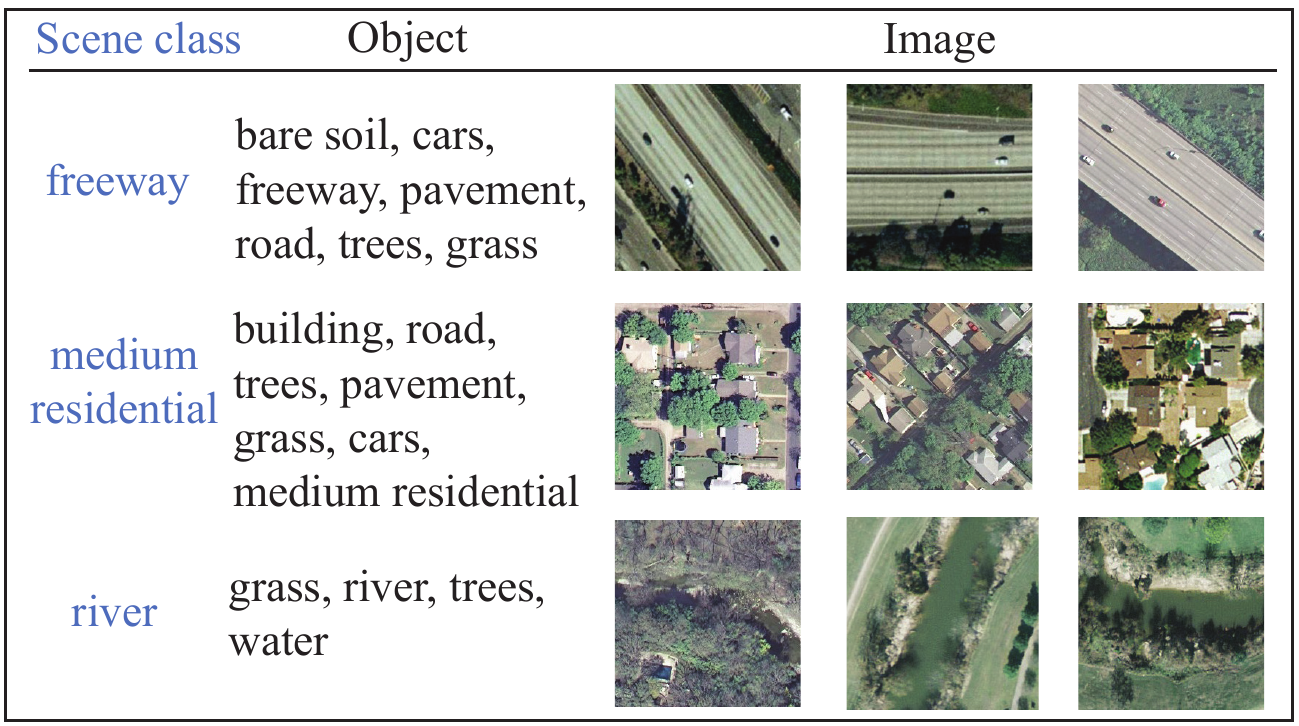} 
	\caption{The unique characteristics of remote sensing images. The first one is object co-occurrence, e.g., ``cars'' and ``road'' tend to appear together in the scene image of ``freeway'', and  ``grass'' and ``water'' jointly occur in the scene class of ``river''. The second one is object spatial correlation, e.g.,  the ``cars'' are usually running or parked on the ``road''. These characteristics can provide more refined descriptions for remote sensing scene images/classes.}
	\label{fig1}
\end{figure}

\underline{F}ew-\underline{S}hot \underline{N}atural \underline{I}mage \underline{C}lassification (FSNIC) problem is closely related to FSRSSC, which aims to quickly recognize novel natural classes from very few examples \cite{YaohuiWang_pr, Zhang_2021_CVPR, SungYZXTH18, tian2021consistent}. The main difference is that the former focuses on natural images while the latter targets at remote sensing scene images. At present, a large number of FSNIC methods have been proposed from the perspective of meta-learning \cite{finn2017model, Zhang_2021_CVPR, wang2021meta}. These methods can be roughly grouped into two categories: (\romannumeral1) metric-based methods that perform novel class prediction via a nearest neighbor classifier with Euclidean \cite{snell2017prototypical}, cosine \cite{ChenLKWH19}, or learnable distance \cite{SungYZXTH18}; and (\romannumeral2) optimization-based methods that learn a new optimization algorithm over few labeled samples to qucikly train a classifier for novel classes \cite{BaikCCKL20, RaghuRBV20, lai2020learning}. Though these methods have shown superior performance on FSNIC, all methods focus on recognizing natural images instead of remote sensing images. Compared to natural images, remote sensing images are more challenging to classify since they have larger intra-class variances and smaller inter-class differences \cite{2020RSmetanet, 2020FewAlajaji, 2020DLAmatchnet, zeng2021prototype, li2021few}. Thus, the classification performance of these FSNIC methods is unsatisfactory when they are directly applied to FSRSSC. 

Recently, several studies \cite{2020FewAlajaji, 2020DLAmatchnet} attempt to extend these existing FSNIC methods to recognize remote sensing scene images. Their basic idea is that designing a good feature extractor \cite{2020FewAlajaji, 2020DLAmatchnet}, metric mechanism \cite{2020RSmetanet}, hyperparameter (e.g., the learning rate) of optimization algorithm \cite{ZhaiLS19} for FSRSSC. Though these FSRSSC methods have achieved promising improvements, they all overlook two intrinsic characteristics of remote sensing images that are different from natural images: (\romannumeral1) {\bf object co-occurrence} refers to that the remote sensing scene images usually contain multiple objects and they usually appear together in a scene image. For example, as shown in Fig.~\ref{fig1}, the objects ``road'' and ``car'' tend to occur together in the scene image of ``free way'', and ``grass'' and ``water'' often accompany the scene class of ``river''. In fact, the object co-occurrence has been investigated in the domain of multi-label remote sensing image classification \cite{qi2020mlrsnet, ji2020multi}, but as far as we know, there is no previous work has explored it on FSRSSC; and (\romannumeral2) {\bf object spatial correlation} refers to that these co-occurrence objects are distributed in the scene image following some spatial structure patterns, which means that these objects may be correlated with each other. For instance, as shown in Fig.~\ref{fig1}, the medium residential scene can be described with the spatial structure of co-occurrence objects: 1) the ``tree'' usually lines both sides of the ``road'', 2) the ``car'' is running or parked on the ``road'', 3) the ``building'' is usually surrounded by the ``roads'', and 4) a large amount of ``grass'' is distributed in near the ``building''. Such unique characteristics are very beneficial for addressing the FSRSSC problem, which can effectively alleviate the data scarcity issue because they can provide more refined descriptions for remote sensing scene classes. 

To fully leverage these two characteristics (i.e., object co-occurrence and object spatial correlation) of remote sensing images for FSRSSC, in this paper, we propose a scene graph matching-based meta-learning framework, called SGMNet. Our key idea is that encoding each test remote sensing image or each scene class as a scene graph and then evaluating their similarity in a scene graph matching manner. The advantage of such scene graph-based design is that it can provide more spatial structural information for evaluating the similarity between each test remote sensing image and each scene class, since the scene graph nodes can effectively reflect these co-occurrence objects meanwhile the spatial correlation of these objects can be explicitly captured by the scene graph edges. 

To this end, we first pretrain a fully convolutional network on the base scene classes to obtain a good spatial representation for each remote sensing image. Second, given a FSRSSC task, we obtain the spatial representation of each scene class by averaging the spatial representation of all labeled samples from the class. Then, a scene graph matching network is designed to evaluate the similarity score between each test image and each scene class, which consists of a graph construction module and a graph matching module. Here, the former is in charge of mapping the spatial representation of each test image or each scene class to a scene graph by regarding its local features as nodes and the relation between all local features as edges. The latter accounts for evaluating their similarity score at the scene graph level. Finally, based on the similarity scores, we perform the prediction of novel scene classes via a nearest neighbor classifier. 

Our main contributions can be summarized as follows:

\begin{itemize}
	\item We identify two significant characteristics of remote sensing images for FSRSSC, i.e., object co-occurrence and object spatial correlation, which are ignored in the existing FSRSSC methods. These characteristics can provide more refined descriptions for each scene class.
	
	\item To fully utilize these characteristics, we propose a novel scene graph matching-based meta-learning framework for FSRSSC. In the framework, more spatial structural information (co-occurrence objects and their spatial relations) can be explicitly captured for encoding each image or each scene class. This effectively alleviates the data scarcity issue. To our best knowledge, this is first work to explore graph matching mechanism for FSRSSC. 
	
	\item We conduct comprehensive experiments on four real-world datasets. The experimental results demonstrate that the proposed method achieves superior performance over various state-of-the-art approaches.
\end{itemize}

The remaining of this paper is organized as follows: In Section~\ref{section_2}, we discuss some related works on remote sensing scene classification, few-shot natural image classification, few-shot remote sensing scene classification, and graph matching techniques. In Section~\ref{section_3}, we describe the proposed SGMNet framework in detail, including an overall framework, a graph construction module, and a graph matching module. Section~\ref{section_4} shows and analyzes the experimental results on four real-world remote sensing scene classification datasets. Finally, the conclusion is presented in Section~\ref{sec_conclusion}.

\section{Related Work}
\label{section_2}

This paper is related to four research domains, including remote sensing scene classification, few-shot natural image classification, few-shot remote sensing scene classification, and graph matching techniques, respectively. Next, we review them in detail respectively.

\subsection{Remote Sensing Scene Classification}
Remote sensing scene classification (RSSC) aims to recognize each remote sensing image with a scene semantic class, which has gained wide attention in the past few decades \cite{HeFLPP20, wang2020looking}. In the early stage, most existing  methods rely on human-engineering features (e.g., scale-invariant feature transformation (SIFT) \cite{TangLX19}, texture descriptors (TD) \cite{FarooqJ019}, and bag-of-visual-words (BOVW) \cite{YangN10}) and traditional machine learning methods (e.g., support vector machine (SVM) \cite{ChenWW09}) to perform the scene prediction for each remote sensing image. For example, in \cite{ZhuZZXZ16}, Zhu et al. proposed to combine the SIFT and TD features and then employed an SVM classifier to predict the scene classes. Zhao et al. \cite{ZhaoTH14} proposed a rotation-invariant representation model, incorporated it into the BOVW method, and then performed scene prediction by the SVM. 

Though these methods have made some progress in RSSC, their performance improvement depends heavily on the quality of human-engineering features, which limits their performance improvement. To address the drawback, a type of novel RSSC approaches based on deep learning is proposed. This type of methods can automatically learn discriminative features from large amounts of labeled data in an end-to-end train manner. For instance, in \cite{PenattiNS15}, Penatti et al. first introduced a CNN to address the RSSC problem and achieved promising classification performance. Cheng et al. \cite{ChengLYGW17} proposed a novel bag of convolutional features (called BoCF) for RSSC by leveraging the pre-trained CNN features to replace these traditional descriptors. Wang et al. \cite{0009LCL19} explored the attention mechanism and designed a novel attention-based recurrent convolutional network to address the RSSC problem. Wang et al. \cite{wang2020looking} proposed a global-local two-stream architecture to learn multiscale representation for RSSC. Xu et al. \cite{xu2021deep} developed a graph convolution-based deep feature aggregation framework to produce more refined features for RSSC. Though these RSSC methods have shown superior performance, they generally count on massive annotated data. This greatly limits their application scenarios. Different from these studies, our work 1) focuses on the FSRSSC problem where only few labeled samples are available for each scene class and 2) proposes a novel scene graph-based metric strategy to evaluate the similarity score between each remote sensing image and each scene class, which fully exploits object co-occurrence and spatial correlation of remote sensing images for FSRSSC.

\subsection{Few-Shot Natural Image Classification}
FSNIC aims to learn transferable meta-knowledge from base natural classes with sufficient labeled samples and then leverage it to recognize novel natural classes with few examples. Recently, a large number of FSNIC methods have been proposed, which can be roughly divided into two groups: 
1) Metric-based methods. The idea behind this type of method is learning a good metric space where the few-shot classification task can be addressed by a nearest neighbor classifier with Euclidean \cite{snell2017prototypical}, cosine \cite{ChenLKWH19, jung2020few}, earth mover's distance \cite{ZhangCLS20}, and learnable distance \cite{SungYZXTH18}. For example, Snell et al. \cite{snell2017prototypical} averaged the features of all labeled samples to represent each class prototype and then assigned each test sample to the labels of its nearest class prototype. Chen \emph{et al.} \cite{chen2020new} proposed a new baseline for FSNIC by pretraining a feature extractor on entire base classes and then performing class prediction by a cosine nearest-centroid classifier with mean-based prototypes.
2) Optimization-based methods. This line of methods attempts to model an optimization algorithm over few labeled samples under the meta-learning framework \cite{FlennerhagRPVYH20, RajeswaranFKL19, lai2020learning}, aiming to adapt to novel natural classes by a few gradient updates. This goal is achieved by meta-learning the hyperparameters of conventional optimization algorithm such as model initialization \cite{finn2017model}, update rule \cite{RaviL17}, learning rate \cite{LiZCL17}, or weight decay \cite{BaikCCKL20}. For example, Chelsea \emph{et al.} \cite{finn2017model} proposed a two-loop optimization framework where the inner loop is in charge of fine-tuning an initial classifier to recognize novel natural classes by few gradient updates, and the outer loop accounts for improving the generalization ability of the initial classifier. 
Different from these existing methods, our work 1) focuses on recognizing remote sensing images instead of natural images, which have larger intra-class variances and smaller inter-class differences; and 2) proposes a novel scene graph matching-based meta-learning framework for FSRSSC.

\subsection{Few-Shot Remote Sensing Scene Classification}
Few-Shot Remote Sensing Scene Classification (FSRSSC) is closely related to FSNIC, which aims to recognize novel scene classes with few examples. Recent works attempt to address the FSRSSC problem by following the idea of FSNIC. Similarly, these methods can also be roughly divided into two groups: 
1) Metric-based methods. Some recent works extended the metric-based meta-learning method to address FSRSSC problem \cite{2020RSmetanet, 2020FewAlajaji, cheng2021spnet, 2020DLAmatchnet, zeng2021prototype, li2021few}. For example, in \cite{2020FewAlajaji}, Alajaji \emph{et al.} extended prototypical network \cite{snell2017prototypical} by introducing a pre-training strategy on all base scene classes and performing novel scene class prediction by a cosine nearest-centroid classifier. Li \emph{et al.} \cite{2020DLAmatchnet} incorporated channel attention and spatial attention modules with the feature network and proposed feature fusion schemes to achieve discriminative features for FSRSSC. In \cite{2020RSmetanet}, a new learnable metric module and a novel loss function were developed for FSRSSC by combining task-level and sample-level classification loss. 2) Optimization-based methods. In \cite{2020FewShot}, the authors extended the MAML method proposed by \cite{finn2017model} to address the FSRSSC problem and verified its effectiveness. Besides, in \cite{RusswurmW0L20}, Ru{\ss}wurm et al. also extend MAML, but they focus on addressing the cross-cities few-shot land cover classification problem. Their goal is learning meta-knowledge from many cities with abundant data and then transferring to many cities lacking data to recognize the land cover. Different from these methods, our work explores the unique characteristics of remote sensing images, i.e., object co-occurrence and object spatial correlation, by viewing each remote sensing image or each scene class as a scene graph, and addresses it in a scene graph matching manner.

\begin{figure*}[!t]
	\centering
	\includegraphics[width=1.0\textwidth]{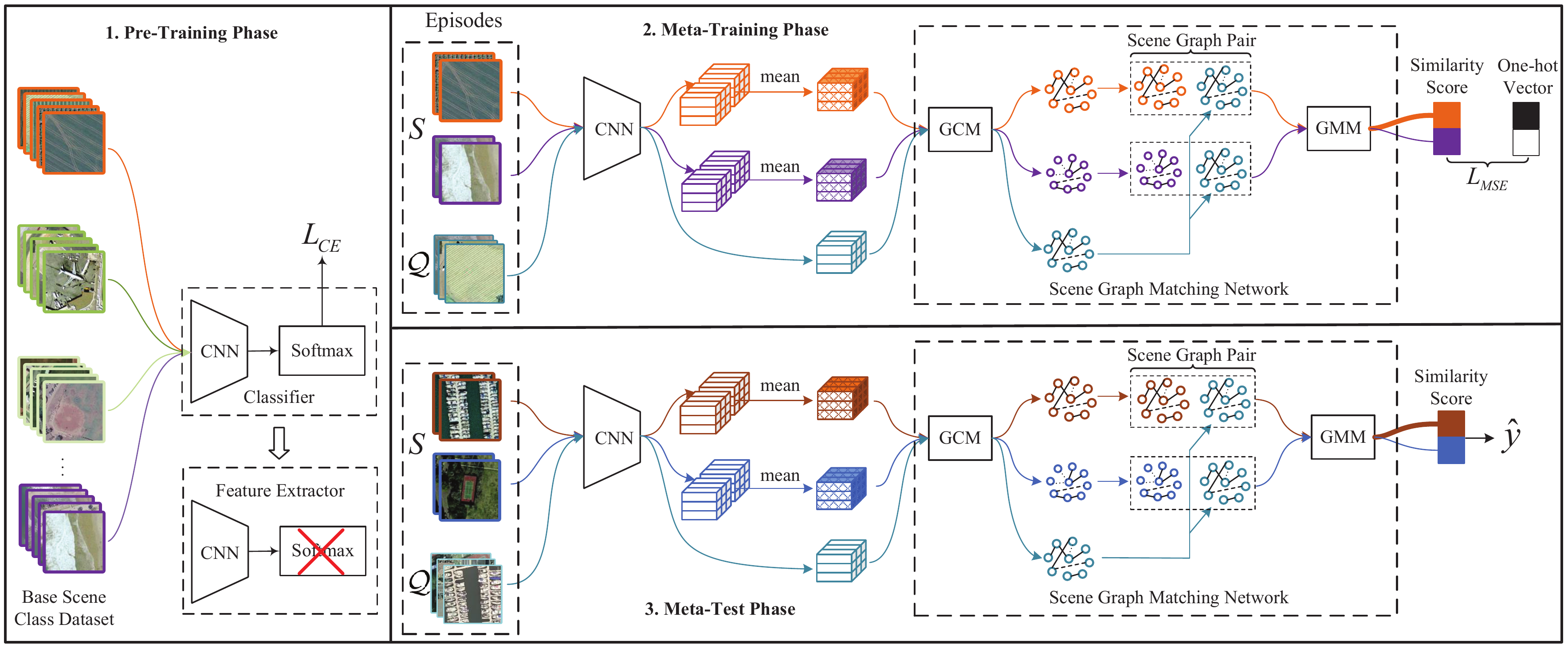} 
	\caption{The scene graph matching-based meta-learning framework for FSRSSC, including three phases: (1) Pre-Training phase that learns a fully convolutional network-based feature extractor on the base scene classes (Section~\ref{section_3_2_1}); (2) Meta-Training phase that learns a scene graph-level metric strategy in an episodic training manner (Section~\ref{section_3_2_3}); and (3) Meta-Test phase that performs novel class prediction by constructing and matching scene graphs (Section~\ref{section_3_2_3}).}
	\label{fig2}
\end{figure*}

\subsection{Graph Matching}
Graph matching aims to evaluate the similarity between two graphs by encoding their structural information including graph node features and complex node relationships. Typically, graph matching can be formulated as a quadratic assignment problem \cite{RiffiS19, LoiolaANHQ07}, i.e., assigning each node in the first graph to a unique node in the second graph \cite{ZanfirS18, YanYH20}. This is a typical NP-hard problem. Thus, traditional graph matching, such as random walk-based methods \cite{ChoLL10} and spectral matching-based methods \cite{LeordeanuH05}, mainly address the matching problem from the view of optimization heuristics \cite{GoldR96, SchellewaldS05}. Though these methods have shown promising performance, all of these optimization methods rely on handcrafted features \cite{liu2020deep}, which limits their performance. Recently, deep learning-based graph matching methods have been proposed and shown superior performance on graph matching. For example, in \cite{LiGDVK19}, a data-driven graph matching network was proposed to compute the similarity between two graphs. Sarlin et al. \cite{SarlinDMR20} regarded the graph matching problem as an optimal transport problem, and attempted to represent each graph by leveraging a graph neural network and perform the graph matching by using a sinkhorn algorithm. To our best knowledge, there is no previous work to explore the graph matching mechanism for FSRSSC. In this paper, we fill the gap by regarding each remote sensing image or each scene class as a scene graph and computing their similarity scores in a graph matching-based manner.

\section{Methodology}
\label{section_3}

This paper focuses on addressing the FSRSSC problem, i.e., recognizing novel scene classes with few labeled samples by leveraging base scene classes with abundant labeled samples. In this section, we first formalize the FSRSSC problem, and then introduce the proposed SGMNet framework and its core component, i.e., the scene graph matching network.

\subsection{Problem Definition}
In the FSRSSC problem, three datasets are given: a base scene dataset $\mathcal{D}_{base}$ with abundant labeled samples, a novel scene training dataset $\mathcal{S}$ (called support set) with few labeled samples, and a novel scene test dataset $\mathcal{Q}$ (called query set) consisting of unlabeled samples. In the base scene dataset $\mathcal{D}_{base}$, there is a large number of remote sensing scene images which are labeled with a base scene class $y_i \in \mathcal{C}_{base}$ where $\mathcal{C}_{base}$ is the set of base scene classes. In the support set $\mathcal{S}$, there is $N$ novel scene classes and each scene class only contains $K$ labeled samples for novel scene classes learning. We denote the set of novel scene classes as  $\mathcal{C}_{novel}$. Note that the base scene class set and novel scene class set are disjoint, \emph{i.e.,} $\mathcal{C}_{base} \cap \mathcal{C}_{novel} = \emptyset$. In the query set $\mathcal{Q}$, there are some unlabeled instances sampled from the scene classes $\mathcal{C}_{novel}$. 

Our goal is to learn a good classifier $f_{\theta}()$ with parameters $\theta$ for query set $\mathcal{Q}$ by leveraging the support set $\mathcal{S}$ and the base scene class dataset $\mathcal{D}_{base}$. The problem is called to $N$-way $K$-shot FSRSSC problem, which can be expressed as:
\begin{equation}
\min\limits_{\theta} \ \mathbb{E}_{(x, \ y) \sim {\mathcal{Q}}} -log(P(y|x, \ \theta, \ \mathcal{S}, \ \mathcal{D}_{base})),
\label{eq0}
\end{equation} where $P(y|x, \ \theta, \ \mathcal{S}, \ \mathcal{D}_{base})$ denotes the probability of classifying each test remote sensing scene image $x \in \mathcal{Q}$ to the novel scene class $y \in \mathcal{C}_{novel}$. 

\subsection{SGMNet Framework}
In this section, to fully leverage the unique characteristics of remote sensing images (i.e., object co-occurrence and object spatial correlation), we propose a scen graph matching-based meta-learning framework for FSRSSC, called SGMNet. Our core idea is regarding each remote sensing scene image or each scene class as a scene graph, and then evaluating the similarity of each test remote sensing image and each scene class in a scene graph matching-based manner. The advantage of such design is that these co-occurrence objects and their spatial correlations can be explicitly captured via the scene graph nodes and their edges, thereby producing more accurate similarity evaluation. As shown in Fig.~\ref{fig2}, the SGMNet framework consists of three phases: pre-training, meta-training, and meta-test phases. 
Next, we introduce them in details, respectively.

\subsubsection{Pre-Training}
\label{section_3_2_1}
We first construct a fully convolutional neural network (CNN), which consists of a feature extractor $f_{\theta_f}()$ and a softmax-based classification head $f_{\theta_c}()$. Then, we train the above network by minimizing the negative loglikelihood estimation on base scene datasets $\mathcal{D}_{base}$. That is,
\begin{equation}
\min\limits_{\{\theta_f, \theta_c\}} \ \mathbb{E}_{(x, \ y) \sim \mathcal{D}_{base}} -log(P(y|x, \ \theta_f, \ \theta_c)),
\label{eq1}
\end{equation} where $\theta_f$ and $\theta_c$ denote the parameters of the feature extractor $f_{\theta_f}()$ and the softmax-based classification head $f_{\theta_c}()$, respetively. Following \cite{RodriguezLDL20}, we also add a self-supervised loss, i.e., the rotation ($0^{\circ}$, $90^{\circ}$, $180^{\circ}$, and $270^{\circ}$) classification loss, aiming to learn a rotation invariant representation for each remote sensing image. Finally, we remove the softmax-based classification head $f_{\theta_c}()$ and the last global pooling layer of feature extractor $f_{\theta_f}()$, aiming to obtain a good spatial representation for each image. Note that the feature extractor $f_{\theta_f}()$ is frozed in the later meta-training and meta-test phases.

\begin{figure*}[!t]
	\centering
	\includegraphics[width=1.0\textwidth]{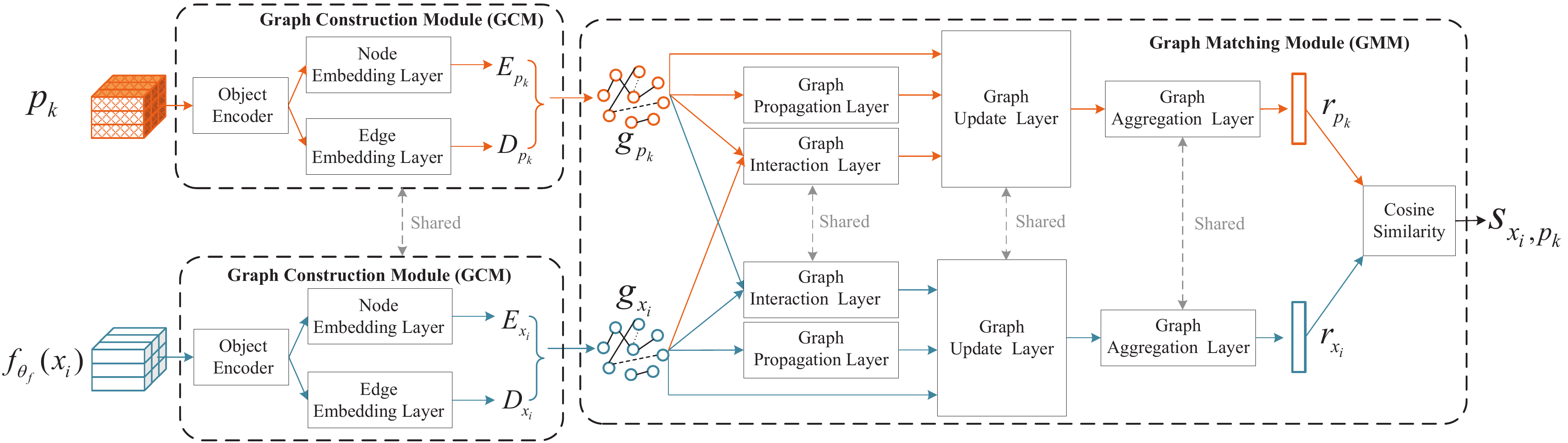} 
	\caption{The illustration of scene graph matching network, including GCM and GMM. The GCM aims to represent each image as a scene graph. The GMM accounts for computing the similarity score of each remote sensing image and each scene class. Note that the parameters of the GCM and GMM are shared. }
	\label{fig3}
\end{figure*}

\subsubsection{Meta-Training}
\label{section_3_2_2}
In this phase, we design a scene graph matching network $f_{\theta_m}()$ with parameters $\theta_m$ as a meta-learner, targeting at representing each remote sensing image or each scene class as a scene graph and then evaluating their similarity in a scene graph matching manner. Please refer to Section~\ref{section_3_3} for the details of meta-learner. Here, we mainly focus on introducing the workflow of the meta-training phase, i.e., how to train the meta-learner $f_{\theta_m}()$ to perform few-shot scene classification tasks in a scene graph matching-based manner.  

As shown in Fig.~\ref{fig2}, we mimic the $N$-way $K$-shot task setting and construct a large set of few-shot remote sensing classification tasks (called episodes) from the base scene dataset $\mathcal{D}_{base}$. Specifically, for each episode, we randomly select $N$ scene classes from the set of base scene classes $\mathcal{C}_{base}$, sample $K$ remote sensing images per scene class from the base scene dataset $\mathcal{D}_{base}$ as support set $\mathcal{S}$, and sample $M$ remote sensing images per scene class as query set $\mathcal{Q}$. Then, we train the meta-learner $f_{\theta_m}()$ in an episodic training manner \cite{vinyals2016matching}. 

Specifically, for each few-shot remote sensing classification task, we first leverage the pre-trained feature extractor $f_{\theta_f}()$ to obtain a good spatial representation $f_{\theta_f} (x_i)$ for each remote sensing image $x_i \in \mathcal{S} \cup \mathcal{Q}$. Second, for each scene class $k$, we estimate its spatial representation $p_k$ in a spatial representation average manner. That is, 
\begin{equation}
	p_k = \frac{1}{|\mathcal{S}_k|} \sum_{x_i \in \mathcal{S}_k} f_{\theta_f} (x_i)
	\label{eq3}
\end{equation}where $\mathcal{S}_k$ denotes the remote sensing image set extracted from scene class $k$ and $|\cdot|$ denotes the size of a set. Third,
we feed the spatial representation $f_{\theta_f} (x_i)$ (or $p_k$) of each query sample $x_i \in \mathcal{Q}$ (or each scene class $k$) into a \underline{G}raph \underline{C}onstruction \underline{M}odule (GCM) $f_{\theta_{sc}}()$, aiming to encode the spatial representation as a scene graph $g_{x_i}$ (or $g_{k}$). This is, 
\begin{equation}
\begin{aligned}
g_{x_i} &=\ f_{\theta_{sc}}(f_{\theta_f} (x_i)), \\
g_{p_k} &=\ f_{\theta_{sc}}(p_i),
\end{aligned}
\label{eq4}
\end{equation}
where $\theta_{sc}$ denotes the parameters of the GCM $f_{\theta_{sc}}()$. Fourth, we evaluate the similarity score $s_{i, k}$ between each query sample $x_i \in \mathcal{Q}$ and each scene class $k$ by leveraging a \underline{G}raph \underline{M}atching \underline{M}odule (GMM) $f_{\theta_{sm}}()$. That is,
\begin{equation}
s_{x_i, p_k} =\ f_{\theta_{sm}}(g_{x_i}, g_{p_k}),\ s_{x_i, p_k} \in [0, 1].
\label{eq5}
\end{equation}where $\theta_{sm}$ denotes the parameters of the GMM  $f_{\theta_{sm}}()$.

Finally, following \cite{SungYZXTH18}, we take the Mean-Square Error (MSE) as the loss function to train our meta-optimizer:
\begin{equation}
\min\limits_{\theta_m} L_{MSE} = MSE(s_{x_i, p_k}, y_{x_i,k}),
\label{eq7}
\end{equation}where $y_{i}$ is a one-hot vector converted by the label of sample $x_i$ and $\theta_m$ consists of $\theta_{sc}$ and $\theta_{sm}$, i.e., $\theta_m = \{\theta_{sc}, \theta_{sm}\}$.

\subsubsection{Meta-Test}
\label{section_3_2_3}
In this phase, we focus on how to recognize the novel scene classes with few labeled samples in a scene graph matching-based manner. The workflow is similar to the meta-training phase. The only difference is that we remove the training step described in Eq.~\ref{eq7}. We perform novel scene class prediction by evaluating the similarity between each test sample and all scene classes by following Eqs.~\ref{eq4}\ -\ \ref{eq5} and then assigning it to the label of the most similar scene class.

\subsection{Scene Graph Matching Network}
\label{section_3_3}
In the above SGMNet framework, the key challenge is how to fully leverage these two intrinsic characteristics of remote sensing images, i.e., the object co-occurrence and object spatial correlation, to evaluate the similarity between each test remote sensing image and each scene class. Our idea is designing a scene graph matching network to encode each test remote sensing image or each scene class as a scene graph, where the nodes reflect these co-occurrence objects meanwhile the edges capture the spatial correlations between these co-occurrence objects, and then perform their similarity evaluation in a scene graph matching-based manner.

As shown in Fig.~\ref{fig3}, the scene graph matching network consists of two key components, i.e., a graph construction module (GCM) and a graph matching module (GMM). Specifically, the GCM takes the spatial representations as inputs, aiming to encode each test remote sensing image or each scene class as a scene graph. After that, the GMM takes the pair of scene graphs as inputs, accounting for utilizing the object co-occurrence and object spatial correlation to evaluate their similarity. Next, we introduce them in detail, respectively.

\begin{figure}[!t]
	\centering
	\includegraphics[width=0.50\textwidth]{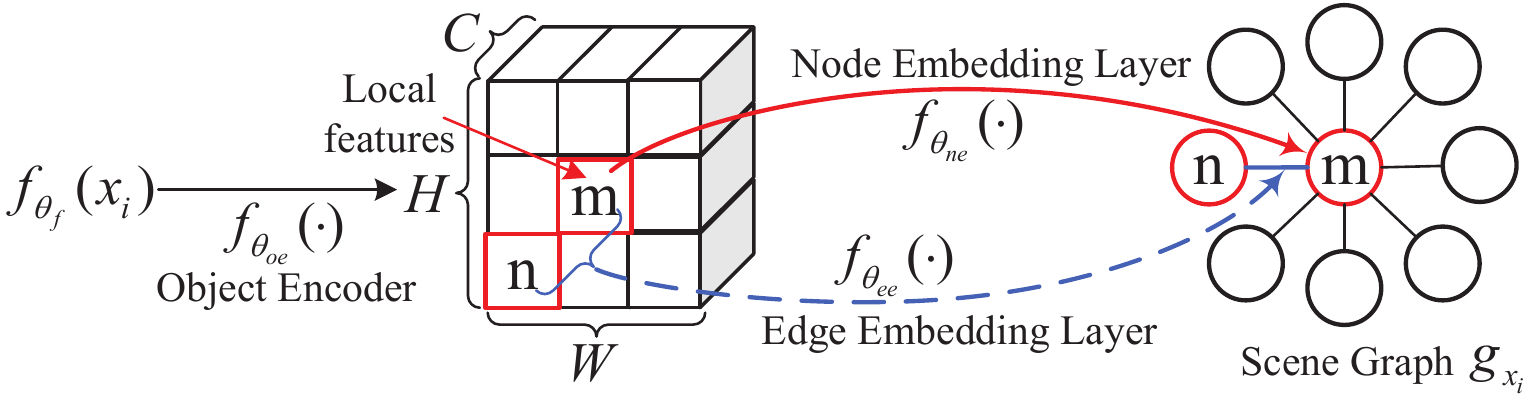} 
	\caption{The illustration of GCM. For each spatial representation $f_{\theta_f} (x_i)$, we first employ an object encoder to capture co-occurrence objects, then represent its local features as the nodes and its relations between all local features as the edges. Here, we only take nodes $m$ and $n$ as an example to show GCM. }
	\label{fig3_1}
\end{figure}

\subsubsection{Graph Construction Module (GCM)}
To fully exploit the characteristics of remote sensing images, i.e., the object co-occurrence and object spatial correlation, we propose to encode each test remote sensing image $x_i \in \mathcal{Q}$ or each scene class $k$ as a scene graph. Our notion is treating their spatial representation $f_{\theta_f} (x_i)$ or $p_k$ as inputs and then constructing a scene graph by regarding their local features as nodes and their relations between all local features as edges. The advantage of such scene graph-based design is that some co-occurrence objects and their spatial structure can be explicitly captured by its graph nodes and edges, respectively, for each remote sensing image or each scene class. Next, for clarity, we take the spatial representation $f_{\theta_f} (x_i)$ of image $x_i \in \mathcal{Q}$ as an example to introduce how to construct the scene graph. The construction process for the scene class $k$ is similar. 

As shown in Fig.~\ref{fig3_1}, given a spatial representation $f_{\theta_f} (x_i)$ of each test sample $x_i \in \mathcal{Q}$, we first encode it to a new spatial representation by an object encoder $f_{\theta_{oe}}()$, aiming to capture these co-occurrence objects from the spatial representation $f_{\theta_f} (x_i)$. The new spatial representation will be a $W \times H \times C$ tensor, which can be considered as a set of
$M$ ($M=WH$) $C$-dimensional local features as $\{l_{x_i, m}\}_{m=0}^{M-1}$. Then, we employ a node embedding layer $f_{\theta_{ne}}()$ to encode these local features $\{l_{x_i, m}\}_{m=0}^{M-1}$ as the node embeddings $E_{x_i}=\{e_{x_i, m}\}_{m=0}^{M-1}$, which would denote potential objects. Besides, an edge embedding layer $f_{\theta_{ee}}()$ is leveraged to explore its pair-wise relationship between two local features $l_{x_i, m}$ and $l_{x_i, n}$ as the edge embeddings $D_{x_i}=\{d_{x_i,m,n}\}_{m,n=0}^{M-1}$ where $m=0,1,...,M-1$ and $n=0,1,...,M-1$. Finally, the above encoding process is formally expressed as:
\begin{equation} 
	\begin{aligned}
		\{l_{x_i, m}\}_{m=0}^{M-1} &= f_{\theta_{oe}}(f_{\theta_f} (x_i)),\\
		\{e_{x_i, m}\}_{m=0}^{M-1} &= f_{\theta_{ne}}(\{l_{x_i, m}\}_{m=0}^{M-1}),
		\\ \{d_{x_i,m,n}\}_{m,n=0}^{M-1} = f_{\theta_{ee}}&(\{l_{x_i, m}||\hat{m}||l_{x_i, n}||\hat{n}\}_{m=0,n=0}^{M-1}),
	\end{aligned}
	\label{eq16_1}
\end{equation}where $\cdot||\cdot$ denote a concatenation operator of two vectors, and $\hat{m}$ and $\hat{n}$ denotes the one-hot vector convered by the position number $m$ and $n$, respectively. In particular, the goal of introducing $\hat{m}$ and $\hat{n}$ as the inputs of edge embedding layer $f_{\theta_{ee}}()$ is to model the spatial structure pattern.

As a result, we obtain a new scene graph-based representation for each test sample $x_i \in \mathcal{Q}$, i.e., $g_{x_i} = <E_{x_i}, D_{x_i}>$. Similarly, for each scene class $k$, its scene graph-based representation $g_{p_k} = <E_{p_k}, D_{p_k}>$ can also be obtained given its spatial representation $p_k$, where $ E_{p_k}=\{e_{p_k,m}\}_{m=0}^{M-1}$ and $D_{p_k}=\{d_{p_k,m,n}\}_{m,n=0}^{M-1}$. 

\subsubsection{Graph Matching Module (GMM)}
Inspired by recent studies \cite{LiGDVK19, SarlinDMR20, ling2020multi}, which evaluate the similarity of graph-structured objects from the perspective of graph matching, we propose a graph matching module to compute the similarity score between the pair of scene graphs (i.e., $g_{x_i}$ and $g_k$). To our best knowledge, there is no work to explore the graph matching mechanism for FSRSSC. In this paper, we fill the gap by encoding each remote sensing image or each scene class as a scene graph and then evaluating their similarity in a scene graph matching-based manner. Specifically, as shown in Fig.~\ref{fig3}, the GMM consists of a graph propagation layer $f_{\theta_{gp}}()$ with parameters $\theta_{gp}$, a graph interaction layer $f_{\theta_{gi}}()$ with parameters $\theta_{gi}$, a graph update layer $f_{\theta_{gu}}()$ with parameters $\theta_{gu}$, and a graph aggregation layer $f_{\theta_{ga}}()$ with parameters $\theta_{ga}$. 
The idea of such design is that learning a more robust representation for each test remote sensing image or each scene class by fully utilizing their intra-graph and cross-graph information and then evaluate their similarity via a cosine similarity function. Next, we elaborate on them, respectively.

{\bf Graph Propagation Layer.} To fully exploit the intra-graph information (i.e., the co-occurrence objects and their spatial correlation) of each scene graph $g_{x_i} = <E_{x_i}, \mathcal{D}_{x_i}>$, we compute a intra-graph node representation $e^{intra}_{x_i,m}$ for each node $m$ by propagating its all neighborhood node embeddings along the edge embeddings $\mathcal{D}_{x_i}$ of the scene graph $g_{x_i}$. Specifically, for each node $m \in [0, M-1]$, we first concatenate its node embedding $e_{x_i, m}$, the embedding $e_{x_i,n}$ of each neighborhood node $n \in [0, M-1]$, and their edge embedding $d_{x_i, m,n}$, followed by a graph propagation layers $f_{\theta_{gp}}()$ that propagate the node information on the scene graph $g_{x_i}$. Then, we aggregate all neighborhood information as the intra-graph node representation $e^{intra}_{x_i,m}$ in an average manner. That is, 
\begin{equation} 
\begin{aligned}
e^{intra}_{x_i,m} = \frac{1}{M}\sum_{n \in [0, M-1]}f_{\theta_{gp}}(e_{x_i,m}||e_{x_i,n}||d_{x_i,m,n}),
\end{aligned}
\label{eq9}
\end{equation}where $\cdot||\cdot$ denote a concatenation operator. Similarly, we can also obtain a new intra-graph node representation $e^{intra}_{p_k,m}$ for the scene graph $g_{p_k} = <E_{p_k}, \mathcal{D}_{p_k}>$ of each class $k$.

{\bf Graph Interaction Layer.} To fully utilize the cross-graph information, i.e., measuring how well a node in one scene graph (e.g., $g_{x_i}$) can be matched to one or more nodes in the other scene graph (e.g., $g_{p_k}$), we calculate a cross-graph node representation $e^{cross}_{x_i,m}$ for each node $m$ by introducing a graph interaction layer. 
Specifically, we first regard these two scene graphs $g_{x_i}$ and $g_{p_k}$ as inputs of the graph interaction layer. Then, we compute the similarity score of cross-graph nodes by calculating the inner product of their node embedding $e_{x_i,m}$ and $e_{p_k,n}$ where $m,n \in [0, M-1]$. Finally, a new cross-graph node representation $e^{cross}_{x_i,m}$ and $e^{cross}_{p_k,n}$ can be obtained by regarding the similarity score as matching weights, respectively. That is,
\begin{equation} 
\begin{aligned}
e^{cross}_{x_i,m} = & \sum_{n} e_{p_k,n} \frac{{exp}({e_{x_i,m} \cdot e_{p_k,n}})}{\sum_{n} {exp}({e_{x_i,m} \cdot e_{p_k,n})}},
\\e^{cross}_{p_k,n} = & \sum_{m} e_{x_i,m} \frac{{exp}({e_{x_i,m} \cdot e_{p_k,n}})}{\sum_{m} {exp}({e_{x_i,m} \cdot e_{p_k,n})}}.
\end{aligned}
\label{eq11}
\end{equation}

{\bf Graph Update Layer.} Till now, we have obtained three types of node representations for the scene graph $g_{x_i}$ or $g_{p_k}$, i.e., $e_{x_i,m}$, $e^{intra}_{x_i,m}$ and $e^{cross}_{x_i,m}$, or $e_{p_k,n}$, $e^{intra}_{p_k,n}$ and $e^{cross}_{p_k,n}$. Then, we introduce a graph update layer $f_{\theta_{gu}}()$ to combine these node representation for the scene graph $g_{x_i}$ or $g_{p_k}$. That is,
\begin{equation} 
\begin{aligned}
e^{update}_{x_i,m} = &f_{\theta_{gu}}(e_{x_i,m}||e^{intra}_{x_i,m}||e^{cross}_{x_i,m}),
\\e^{update}_{p_k,n} = & f_{\theta_{gu}}(e_{p_k,n}||e^{intra}_{p_k,n}||e^{cross}_{p_k,n}),
\end{aligned}
\label{eq13}
\end{equation}where $e^{update}_{x_i,m}$ and $e^{update}_{p_k,n}$ denote the combined node representation of the scene graph $g_{x_i}$ and $g_{p_k}$, respectively.

{\bf Graph Aggregation Layer.} To evaluate the graph-level matching score, we encode the entire scene graph $g_{x_i}$ or $g_k$ as a graph-level representation $r_{x_i}$ or $r_k$ by aggregating their node-level representations. That is,

\begin{equation} 
	\begin{aligned}
		r_{x_i} = &\sum_{m=0}^{M-1} f_{\theta_{ga}}(e^{update}_{x_i,m}, \frac{1}{M}\sum_{m=0}^{M-1} e^{update}_{x_i,m}) e^{update}_{x_i,m},
		\\r_{p_k} = &\sum_{n=0}^{M-1} f_{\theta_{ga}}(e^{update}_{p_k,n}, \frac{1}{M}\sum_{n=0}^{M-1} e^{update}_{p_k,n}) e^{update}_{p_k,n},
	\end{aligned}
	\label{eq15}
\end{equation}where $f_{\theta_{ga}}()$ denotes the graph aggregation layer, which is used for estimating the weights of node aggregation.

{\bf Similarity Evaluation.} Finally, we evaluate the similarity score between $g_{x_i}$ and $g_{p_k}$ by calculating the cosine similarity of two graph-level representation $r_{x_i}$ and $r_{p_k}$. That is, 

\begin{equation} 
\begin{aligned}
s_{x_i,p_k} =\frac{cosine\_similarity(r_{x_i}, r_{p_k})+1}{2},
\end{aligned}
\label{eq16}
\end{equation}where $cosine\_similarity(\cdot)$ denotes the function of cosine similarity between two vectors and $s_{x_i,p_k}$ is scaled to $[0, 1]$.

\section{Experiments}
\label{section_4}
In this section, we evaluate the performance of the proposed SGMNet framework on four real-world datasets, and then discuss the experiment results and present our ablation study.

\subsection{Datasets and Settings}
We conduct the experiments of 5-way 1-shot/5-shot tasks on four real-world datasets, i.e., UCMercedLandUse, NWPU-RESISC45, WHU-RS19, AID, and MLRSNet. The details are shown in Table~\ref{table1}. Next, we elaborate on them, respectively.

\subsubsection{UCMercedLandUse}
The dataset is a classical and common dataset for evaluating the performance of RSSC methods. It is composed of 21 scene classes: agricultural, airplane, baseball diamond, beach, buildings, chaparral, dense residential, forest, freeway, golf course, harbour, intersection, medium density residential, mobile home park, overpass, parking lot, river, runway, sparse residential, storage tanks, and tennis courts. For each scene class, there are 100 remote sensing images with a spatial resolution of 256$\times$256. Following \cite{2020DLAmatchnet}, we split the dataset into 10 scene classes for meta-training, 5 scene classes for meta-validation, and 6 scene classes for meta-test, respectively.

\subsubsection{WHU-RS19}
The dataset is proposed by Wuhan University for RSSC, which involves 1005 remote sensing images with a spatial resolution of 600$\times$ 600. It includes 19 scene classes, including airport, bridge, desert, football field, industrial, mountain, parking, port, residential, beach, farmland, forest, park, railway station, commercial, meadow, pond, river, and viaduct. Following \cite{2020DLAmatchnet}, we split the dataset into 9 scene classes for meta-training, 5 scene classes for meta-validation, and 5 scene classes for meta-test, respectively.

\begin{table}
	\caption{The basic statistics of the UCMercedLandUse, WHU-RS19, NWPU-RESISC45, AID, and MLRSNet datasets.}\smallskip
	\centering
	\smallskip\scalebox
	{1.00}{\begin{tabular}{l|c|c|c}
			\hline
			\multicolumn{1}{c|}{\multirow{2}{*}{Datasets}} & \multicolumn{3}{c}{\multirow{1}{*}{The number of scene classes}} \\
			\cline{2-4}
			& meta-training & meta-validation & meta-test\\
			\hline
			\hline
			UCMercedLandUse & 10 & 5 & 6 \\
			WHU-RS19 & 9 & 5 & 5 \\
			NWPU-RESISC45 & 25 & 10 & 10\\
			AID & 16 & 7 & 7\\
			\hline
	\end{tabular}}
	\label{table1}
\end{table}

\begin{table*}[t]
	\caption{Experiment results on the UCMercedLandUse and WHU-RS19 datasets. Top two results are shown in bold and underline.}\smallskip
	\centering
	\smallskip\scalebox
	{1.0}{\begin{tabular}{l|c|c|c|c|c|c}
			\hline
			\multicolumn{1}{l|}{\multirow{2}{*}{Method}}&
			\multicolumn{1}{c|}{\multirow{2}{*}{Type}}& \multicolumn{1}{c|}{\multirow{2}{*}{Backbone}}&
			\multicolumn{2}{|c|}{UCMerced LandUse} & \multicolumn{2}{|c}{WHU-RS19} \\ 
			\cline{4-7}
			& & & 5-way 1-shot & 5-way 5-shot & 5-way 1-shot & 5-way 5-shot \\
			\hline \hline
			MAML \cite{finn2017model} & Optimization & Conv-5 & 48.86 $\pm$ 0.74$\%$  & 60.78 $\pm$ 0.62$\%$ & 49.13 $\pm$ 0.65$\%$  & 62.49 $\pm$ 0.51$\%$ \\
			MetaSGD \cite{LiZCL17} & Optimization & Conv-5 & 50.52 $\pm$ 2.61$\%$  & 60.82 $\pm$ 2.00$\%$ & 51.54 $\pm$ 2.31$\%$  & 61.74 $\pm$ 2.02$\%$ \\
			LLSR \cite{ZhaiLS19} & Optimization & Conv-5 & 39.47$\%$  & 57.40$\%$ & 57.10$\%$  & 70.65$\%$ \\
			ProtoNet \cite{snell2017prototypical} & Metric & Conv-5 & 52.27 $\pm$ 0.20\%  & 69.86 $\pm$ 0.15$\%$ & 58.01 $\pm$ 0.16\%  & 80.70 $\pm$ 0.11\% \\
			MatchingNet \cite{vinyals2016matching} & Metric & Conv-5 & 34.70$\%$  & 52.71$\%$ & 50.13$\%$  & 54.10$\%$ \\
			RelationNet \cite{SungYZXTH18} & Metric & Conv-5 & 48.08 $\pm$ 1.67$\%$  & 61.88 $\pm$ 0.50$\%$ & 60.92 $\pm$ 1.86$\%$  & 79.75 $\pm$ 1.19$\%$ \\
			DeepEMD \cite{ZhangCLS20} & Metric & Conv-5 & \underline{58.47 $\pm$ 0.76$\%$}  & 70.42 $\pm$ 0.58$\%$ & 63.76 $\pm$ 0.74$\%$  & 78.25 $\pm$ 0.43$\%$ \\
			FCASIM \cite{ZhangBWBL21} & Metric & Conv-5 & 56.63 $\pm$ 0.70$\%$  & \underline{71.30 $\pm$ 0.52$\%$} & 74.40 $\pm$ 0.69$\%$  & 87.10 $\pm$ 0.43$\%$ \\
			RS-MetaNet \cite{2020RSmetanet} & Metric & Conv-5 & $49.68 \pm 0.71\%$  & $67.53 \pm 0.59\%$ & \underline{$74.58 \pm 0.68\%$}  & \underline{$ 87.45 \pm 0.40\%$} \\
			DLA-MatchNet \cite{2020DLAmatchnet} & Metric & Conv-5 & 53.76 $\pm$ 0.62$\%$  & 63.01 $\pm$ 0.51$\%$ & 68.27 $\pm$ 1.83$\%$  & 79.89 $\pm$ 0.33$\%$ \\
			\hline
			Our Method (SGMNet) & Metric & Conv-5 & \textbf{60.52} $\pm$ \textbf{0.74}$\%$ & \textbf{73.42} $\pm$ \textbf{0.49}$\%$ & \textbf{85.06} $\pm$ \textbf{0.55}$\%$ & \textbf{90.12} $\pm$ \textbf{0.27}$\%$ \\
			\hline
	\end{tabular}}
	\label{table2}
\end{table*}

\begin{table*}[t]
	\caption{Experiment results on the AID and NWPU-RESISC45 datasets. Top two results are shown in bold and underline. }\smallskip
	\centering
	\smallskip\scalebox
	{1.0}{\begin{tabular}{l|c|c|c|c|c|c}
			\hline
			\multicolumn{1}{l|}{\multirow{2}{*}{Method}}&
			\multicolumn{1}{c|}{\multirow{2}{*}{Type}}& \multicolumn{1}{c|}{\multirow{2}{*}{Backbone}}&
			\multicolumn{2}{|c|}{AID} & \multicolumn{2}{|c}{NWPU-RESISC45} \\ 
			\cline{4-7}
			& & & 5-way 1-shot & 5-way 5-shot & 5-way 1-shot & 5-way 5-shot \\
			\hline \hline
			MAML \cite{finn2017model} & Optimization & Conv-5 & 43.20 $\pm$ 0.77$\%$  & 60.37 $\pm$ 0.75$\%$ & 48.40 $\pm$ 0.82$\%$  & 62.90 $\pm$ 0.69$\%$ \\
			MetaSGD \cite{LiZCL17} & Optimization & Conv-5 & 45.01 $\pm$ 0.98$\%$  & 62.58 $\pm$ 0.80$\%$ & 60.63 $\pm$ 0.90$\%$  & 75.75 $\pm$ 0.65$\%$ \\
			LLSR \cite{ZhaiLS19} & Optimization & Conv-5 & 45.18$\%$  & 61.76$\%$ & 51.43$\%$  & 72.90$\%$ \\			
			ProtoNet \cite{snell2017prototypical} & Metric & Conv-5 & 54.32 $\pm$ 0.86\%  & 67.80 $\pm$ 0.64\% &  40.33 $\pm$ 0.18\%  & 63.82 $\pm$ 0.56$\%$ \\
			MatchingNet \cite{vinyals2016matching} & Metric & Conv-5 & 33.87$\%$  & 50.40$\%$ & 37.61$\%$  & 47.10$\%$ \\
			RelationNet \cite{SungYZXTH18} & Metric & Conv-5 & 54.62 $\pm$ 0.80$\%$  & 68.80 $\pm$ 0.66$\%$ & 66.43 $\pm$ 0.73$\%$  & 78.35 $\pm$ 0.51$\%$\\
			DeepEMD \cite{ZhangCLS20} & Metric & Conv-5 & 61.04 $\pm$ 0.77$\%$  & 74.51 $\pm$ 0.55$\%$ & 64.39 $\pm$ 0.84$\%$  & 78.01 $\pm$ 0.56$\%$\\
			FCASIM \cite{ZhangBWBL21} & Metric & Conv-5 & 61.37 $\pm$ 0.95$\%$  & 74.47 $\pm$ 0.65$\%$ & 67.58 $\pm$ 0.85$\%$  & 81.56 $\pm$ 0.56$\%$ \\
			DLA-MatchNet \cite{2020DLAmatchnet} & Metric & Conv-5 & \underline{61.99 $\pm$ 0.94$\%$} & \underline{75.03 $\pm$ 0.67$\%$} &  \underline{68.80 $\pm$ 0.70$\%$}  & \underline{81.63 $\pm$ 0.46$\%$} \\
			RS-MetaNet \cite{2020RSmetanet} & Metric & Conv-5 & $58.51 \pm 0.84\%$  & $73.76 \pm 0.69\%$ & $64.07 \pm 0.90\%$  & $ 79.62 \pm 0.65\%$ \\ 
			\hline
			Our Method (SGMNet) & Metric & Conv-5 & \textbf{62.21} $\pm$ \textbf{0.81}$\%$ & \textbf{75.68} $\pm$ \textbf{0.56}$\%$ & \textbf{70.40} $\pm$ \textbf{0.83}$\%$ & \textbf{82.32} $\pm$ \textbf{0.53}$\%$ \\
			\hline
	\end{tabular}}
	\label{table3}
\end{table*}

\subsubsection{AID}
The dataset is another large-scale dataset for RSSC including 30 scene classes: agricultural, airplane, baseball diamond, beach, buildings, chaparral, dense residential, forest, freeway, golf course, harbour, intersection, medium density residential, mobile home park, overpass, parking lot, river, runway, sparse residential, storage tanks, and tennis courts. In the dataset, each scene class contains around 220 $\sim$ 420 images with the size of 600$\times$ 600. Referring to \cite{2020DLAmatchnet}, we split the dataset into 16 scene classes for meta-training, 7 scene classes for meta-validation, and 7 scene classes for meta-test, respectively (see Appendix for more details).

\subsubsection{NWPU-RESISC45}
The dataset is a large-scale RSSC dataset, which consists of 45 scene classes: airplane, airport, baseball diamond, basketball court, beach, bridge, chaparral, church, circular farmland, cloud, commercial area, dense residential, desert, forest, freeway, golf course, ground track field, harbour, industrial area, intersection, island, lake, meadow, medium residential, mobile home park, mountain, overpass, palace, parking lot, railway, railway station, rectangular farmland, river, roundabout, runway, sea ice, ship, snowberg, sparse residential, stadium, storage tank, tennis court, terrace, thermal power station, and wetland. For each scene class, there are 700 remote sensing images with a spatial resolution of 256$\times$256. Following \cite{2020DLAmatchnet}, we split the 45 scene classes into 25 scene classes for meta-training, 10 scene classes for meta-validation, and 10 scene classes for meta-test, respectively.

\subsection{Implementation Details}
\subsubsection{Network Architecture}
Following \cite{2020DLAmatchnet}, we conduct the experiments by using a shallow backbone (i.e., Conv-5) as the feature extractor, which introduces five convolution blocks for representing each remote sensing image. Here, each convolution block consists of a 3$\times$3 convolution layer, a batch normalization layer, and a ReLU-based activation function layer, respectively. In particular, the first four convolution blocks additionally introduce a 2$\times$2 max-pooling layers, while the last one does not. In the scene graph matching network: (\romannumeral1) we employ a convolution block with 256 channels as the object encoder, which produces a new spatial representation with size of $4\times4\times256$ (UCMercedLandUse and NWPU-RESISC45) or $8\times8\times256$ (WHU-RS19 and AID); (\romannumeral2) we employ a two-layers fully connected network with 256 hidden units and 128-dimensional outputs as the node embedding layer; (\romannumeral3) we employ a two-layers fully connected network with 256 hidden units and 64-dimensional outputs, as the edge embedding layer; (\romannumeral4) we employ a two-layers fully connected network with 512 hidden units and 512-dimensional outputs as the graph propagation layer; (\romannumeral5) we employ a single-layers fully connected network with 256-dimensional outputs as the graph update layer; and (\romannumeral6) we employ a single-layers fully connected network as the graph aggregation layer. In the above all modules, ReLU is employed as the activation function.

\subsubsection{Training Details}
In pre-training phase, we pre-train the feature extractor with 100 epochs on the entire base classes by using an SGD optimizer with a momentum of 0.9 and a weight decay of 0.0005. Here, the learning rate is initialized as 0.1 and then decayed by 0.1 at epochs 60, 80, and 90, respectively. In meta-training phase, we construct 10000 episodes and then train the scene graph matching network with 100 epochs by an Adam optimizer with a weight decay of 0.0005 in an episodic training manner. Here, the learning rate is set to 0.00001.

\subsubsection{Evaluation}
We evaluate the proposed method on 600 randomly sampled 5-way 1-shot and 5-way 5-shot classification tasks from the test set and report the mean accuracy together with the 95\% confidence interval. Here, we randomly sample 15 remote sensing images per class as the query set to evaluate the classification performance.

\subsubsection{Baseline Methods}
We select some few-shot natural image classification methods (i.e., MAML \cite{finn2017model}, MetaSGD \cite{LiZCL17}, ProtoNet \cite{snell2017prototypical}, MatchingNet \cite{vinyals2016matching}, RelationNet \cite{SungYZXTH18}), and DeepEMD \cite{ZhangCLS20}, and four state-of-the-art FSRSSC methods (i.e., LLSR \cite{ZhaiLS19}, FCASIM \cite{ZhangBWBL21}, RS-MetaNet \cite{2020RSmetanet}, and DLA-MatchNet \cite{2020DLAmatchnet}) as our baselines. Among these methods, MAML, MetaSGD and LLSR can be regarded as optimization-based meta-learning methods, and ProtoNet, MatchingNet, RN, FCASIM, RS-MetaNet, and DLA-MatchNet can be regarded as metric-based meta-learning methods. 

\begin{table*}[t]
	\caption{Effect of backbone on the UCMerced LandUse and WHU-RS19 datasets. }\smallskip
	\centering
	\smallskip\scalebox
	{1.0}{\begin{tabular}{l|c|c|c|c|c}
			\hline
			\multicolumn{1}{l|}{\multirow{2}{*}{Method}}&
			\multicolumn{1}{c|}{\multirow{2}{*}{Backbone}}&
			\multicolumn{2}{|c|}{UCMerced LandUse} & \multicolumn{2}{|c}{WHU-RS19} \\ 
			\cline{3-6}
			& & 5-way 1-shot & 5-way 5-shot & 5-way 1-shot & 5-way 5-shot  \\
			\hline \hline
			Our Method (SGMNet) & Conv-5 (shallow) & 60.52 $\pm$ 0.74$\%$  & 73.42 $\pm$ 0.49$\%$ & 85.06 $\pm$ 0.55$\%$  & 90.12 $\pm$ 0.27$\%$ \\
			Our Method (SGMNet) & ResNet-12 (deep) & 64.17 $\pm$ 0.75$\%$ & 76.63 $\pm$ 0.59$\%$ & 86.32 $\pm$ 0.54$\%$ & 91.02 $\pm$ 0.30$\%$ \\
			\hline
	\end{tabular}}
	\label{table4}
\end{table*}

\begin{table*}[t]
	\caption{Effect of backbone on the AID and NWPU-RESISC45 datasets. }\smallskip
	\centering
	\smallskip\scalebox
	{1.0}{\begin{tabular}{l|c|c|c|c|c}
			\hline
			\multicolumn{1}{l|}{\multirow{2}{*}{Method}}&
			\multicolumn{1}{c|}{\multirow{2}{*}{Backbone}}&
			\multicolumn{2}{|c|}{AID} & \multicolumn{2}{|c}{NWPU-RESISC45} \\ 
			\cline{3-6}
			& & 5-way 1-shot & 5-way 5-shot & 5-way 1-shot & 5-way 5-shot  \\
			\hline \hline
			Our Method (SGMNet) & Conv-5 (shallow) & 62.21 $\pm$ 0.81$\%$  & 75.68 $\pm$ 0.56$\%$ & 70.40 $\pm$ 0.83$\%$  & 82.32 $\pm$ 0.53$\%$ \\
			Our Method (SGMNet) & ResNet-12 (deep) & 64.32 $\pm$ 0.79$\%$  & 77.98 $\pm$ 0.42$\%$ & 73.01 $\pm$ 0.77$\%$  & 84.52 $\pm$ 0.50$\%$ \\
			\hline
	\end{tabular}}
	\label{table5}
\end{table*}

\begin{table*}[t]
	\caption{Ablation study of the scene graph matching network on the UCMerced LandUse and WHU-RS19 datasets. }\smallskip
	\centering
	\smallskip\scalebox
	{1.0}{\begin{tabular}{l|c|c|c|c|c}
			\hline
			\multicolumn{1}{l|}{\multirow{2}{*}{}}&
			\multicolumn{1}{c|}{\multirow{2}{*}{Setting}}&
			\multicolumn{2}{|c|}{UCMerced LandUse} & \multicolumn{2}{|c}{WHU-RS19} \\ 
			\cline{3-6}
			& & 5-way 1-shot & 5-way 5-shot & 5-way 1-shot & 5-way 5-shot  \\
			\hline \hline
			(\romannumeral1) & + Euclidean distance & 54.51 $\pm$ 0.72$\%$  & 71.43 $\pm$ 0.49$\%$ & 75.82 $\pm$ 0.62$\%$  & 84.60 $\pm$ 0.30$\%$ \\
			(\romannumeral2) & + Cosine distance & 57.38 $\pm$ 0.74$\%$ & 71.96 $\pm$ 0.48$\%$ & 75.52 $\pm$ 0.59$\%$ & 87.64 $\pm$ 0.32$\%$ \\
			(\romannumeral3) & + RelationNet & 52.90 $\pm$ 0.73$\%$ & 68.36 $\pm$ 0.53$\%$ & 79.95 $\pm$ 0.44$\%$ & 88.07 $\pm$ 0.38$\%$ \\
			(\romannumeral4) & + SGMNet & \textbf{60.52} $\pm$ \textbf{0.74}$\%$ & \textbf{73.42} $\pm$ \textbf{0.49}$\%$ & \textbf{85.06} $\pm$ \textbf{0.55}$\%$ & \textbf{90.12} $\pm$ \textbf{0.27}$\%$ \\
			\hline
	\end{tabular}}
	\label{table6}
\end{table*}

\begin{table*}[t]
	\caption{Ablation study of the scene graph matching network on the AID and NWPU-RESISC45 datasets. }\smallskip
	\centering
	\smallskip\scalebox
	{1.0}{\begin{tabular}{l|c|c|c|c|c}
			\hline
			\multicolumn{1}{l|}{\multirow{2}{*}{}}&
			\multicolumn{1}{c|}{\multirow{2}{*}{Setting}}&
			\multicolumn{2}{|c|}{AID} & \multicolumn{2}{|c}{NWPU-RESISC45} \\ 
			\cline{3-6}
			& & 5-way 1-shot & 5-way 5-shot & 5-way 1-shot & 5-way 5-shot  \\
			\hline \hline
			(\romannumeral1) & + Euclidean distance & 54.58 $\pm$ 0.83$\%$  & 71.58 $\pm$ 0.61$\%$ & 63.20 $\pm$ 0.89$\%$  & 78.46 $\pm$ 0.56$\%$ \\
			(\romannumeral2) & + Cosine distance & 58.53 $\pm$ 0.81$\%$  & 72.33 $\pm$ 0.61$\%$ & 66.74 $\pm$ 0.88$\%$  & 80.51 $\pm$ 0.53$\%$ \\
			(\romannumeral3) & + RelationNet & 58.75 $\pm$ 0.84$\%$  & 72.10 $\pm$ 0.67$\%$ & 65.58 $\pm$ 0.83$\%$  & 79.20 $\pm$ 0.60$\%$ \\
			(\romannumeral4) & + SGMNet & \textbf{62.21} $\pm$ \textbf{0.81}$\%$  & \textbf{75.68} $\pm$ \textbf{0.56}$\%$ & \textbf{70.40} $\pm$ \textbf{0.83}$\%$ & \textbf{82.32} $\pm$ \textbf{0.53}$\%$ \\
			\hline
	\end{tabular}}
	\label{table7}
\end{table*}

\subsection{Discussion of Results}
In this subsection, we report the experimental results (5-way 1-shot and 5-way 5-shot tasks) of all baseline methods and the proposed SGMNet on two small-scale datasets (i.e., UCMercedLandUse and WHU-RS19) and two large-scale datasets (i.e., AID and NWPU-RESISC45). These experimental results are shown in Table~\ref{table2} and \ref{table3}. 

\subsubsection{The Results on Small-Scale Dataset}
Table~\ref{table2} shows the experimental results of the baselines and the proposed SGMNet on the UCMercedLandUse and WHU-RS19 datasets. From Table~\ref{table2}, we can see that the proposed SGMNet outperforms all baseline methods, by around 1\% $\sim$ 5\% on the classification accuracy of 5-way 1-shot and 5-shot tasks. This verifies the effectiveness of our method. Specifically, compared with the optimization-based approaches, our method achieves 1\% $\sim$ 27\% higher classification accuracy. Different from these optimization-based methods, our method follows the framework of metric-based few-shot learning and proposes to compute the similarity score in the scene graph level, instead of learning a good initialization or optimization algorithm over few labeled samples. These experimental results show the superiority of the proposed SGMNet method. Different from the metric-based approaches, our method explores the object co-occurrence and object spatial correlation by regarding each remote sensing image or each scene class as a scene graph. The experimental results demonstrate that the proposed SGMNet is effective, with an improvement of 1\% $\sim$ 10\%. Finally, we would like to emphasize that compared with these state-of-the-art FSRSSC methods (i.e., LLSR, FCASIM, RS-MetaNet, and DLA-MatchNet), our method outperforms them by a large margin, around 10\% $\sim$ 15\%. The reasons may be that our method (\romannumeral1) regards each remote sensing scene image or each scene class as a scene graph; and (\romannumeral2) evaluates their similarity scores in a scene graph matching-based manner, which fully leverages the object co-occurrence and object spatial correlation of remote sensing images. 
As we analyze in the Fig.~\ref{fig1}, the object co-occurrence and spatial correlation are the unique advantage of remote sensing scene images.

\begin{table*}[t]
	\caption{Ablation study of the scene graph matching network on the UCMerced LandUse and WHU-RS19 datasets. }\smallskip
	\centering
	\smallskip\scalebox
	{1.0}{\begin{tabular}{l|c|c|c|c|c}
			\hline
			\multicolumn{1}{l|}{\multirow{2}{*}{}}&
			\multicolumn{1}{c|}{\multirow{2}{*}{Setting}}&
			\multicolumn{2}{|c|}{UCMerced LandUse} & \multicolumn{2}{|c}{WHU-RS19} \\ 
			\cline{3-6}
			& & 5-way 1-shot & 5-way 5-shot & 5-way 1-shot & 5-way 5-shot  \\
			\hline \hline
			(\romannumeral1) & Our method (SGMNet) & \textbf{60.52} $\pm$ \textbf{0.74}$\%$ & \textbf{73.42} $\pm$ \textbf{0.49}$\%$ & \textbf{85.06} $\pm$ \textbf{0.55}$\%$ & \textbf{90.12} $\pm$ \textbf{0.27}$\%$ \\
			(\romannumeral2) & w/o graph propagation layer & 59.02 $\pm$ 0.71$\%$  & 72.69 $\pm$ 0.48$\%$ & 84.13 $\pm$ 0.57$\%$  & 89.83 $\pm$ 0.26$\%$ \\
			(\romannumeral3) & w/o graph interaction layer & 58.85 $\pm$ 0.71$\%$ & 71.38 $\pm$ 0.46$\%$ & 83.40 $\pm$ 0.53$\%$ & 88.40 $\pm$ 0.27$\%$ \\
			\hline
	\end{tabular}}
	\label{table8}
\end{table*}

\begin{table*}[t]
	\caption{Ablation study of the scene graph matching network on the AID and NWPU-RESISC45 datasets. }\smallskip
	\centering
	\smallskip\scalebox
	{1.0}{\begin{tabular}{l|c|c|c|c|c}
			\hline
			\multicolumn{1}{l|}{\multirow{2}{*}{}}&
			\multicolumn{1}{c|}{\multirow{2}{*}{Setting}}&
			\multicolumn{2}{|c|}{AID} & \multicolumn{2}{|c}{NWPU-RESISC45} \\ 
			\cline{3-6}
			& & 5-way 1-shot & 5-way 5-shot & 5-way 1-shot & 5-way 5-shot  \\
			\hline \hline
			(\romannumeral1) & Our method (SGMNet) &  \textbf{62.21} $\pm$ \textbf{0.81}$\%$  & \textbf{75.68} $\pm$ \textbf{0.56}$\%$ & \textbf{70.40} $\pm$ \textbf{0.83}$\%$ & \textbf{82.32} $\pm$ \textbf{0.53}$\%$ \\
			(\romannumeral2) & w/o graph propagation layer & 61.53 $\pm$ 0.82$\%$  & 74.84 $\pm$ 0.60$\%$ & 67.85 $\pm$ 0.84$\%$  & 81.39 $\pm$ 0.55$\%$ \\
			(\romannumeral3) & w/o graph interaction layer & 61.22 $\pm$ 0.81$\%$  & 73.16 $\pm$ 0.61$\%$ & 67.46 $\pm$ 0.79$\%$  & 81.21 $\pm$ 0.56$\%$ \\
			\hline
	\end{tabular}}
	\label{table9}
\end{table*}

\begin{figure*}
	\centering
	\subfigure[UCMerced LandUse]{ 
		\label{fig4a} 
		\includegraphics[width=0.48\columnwidth]{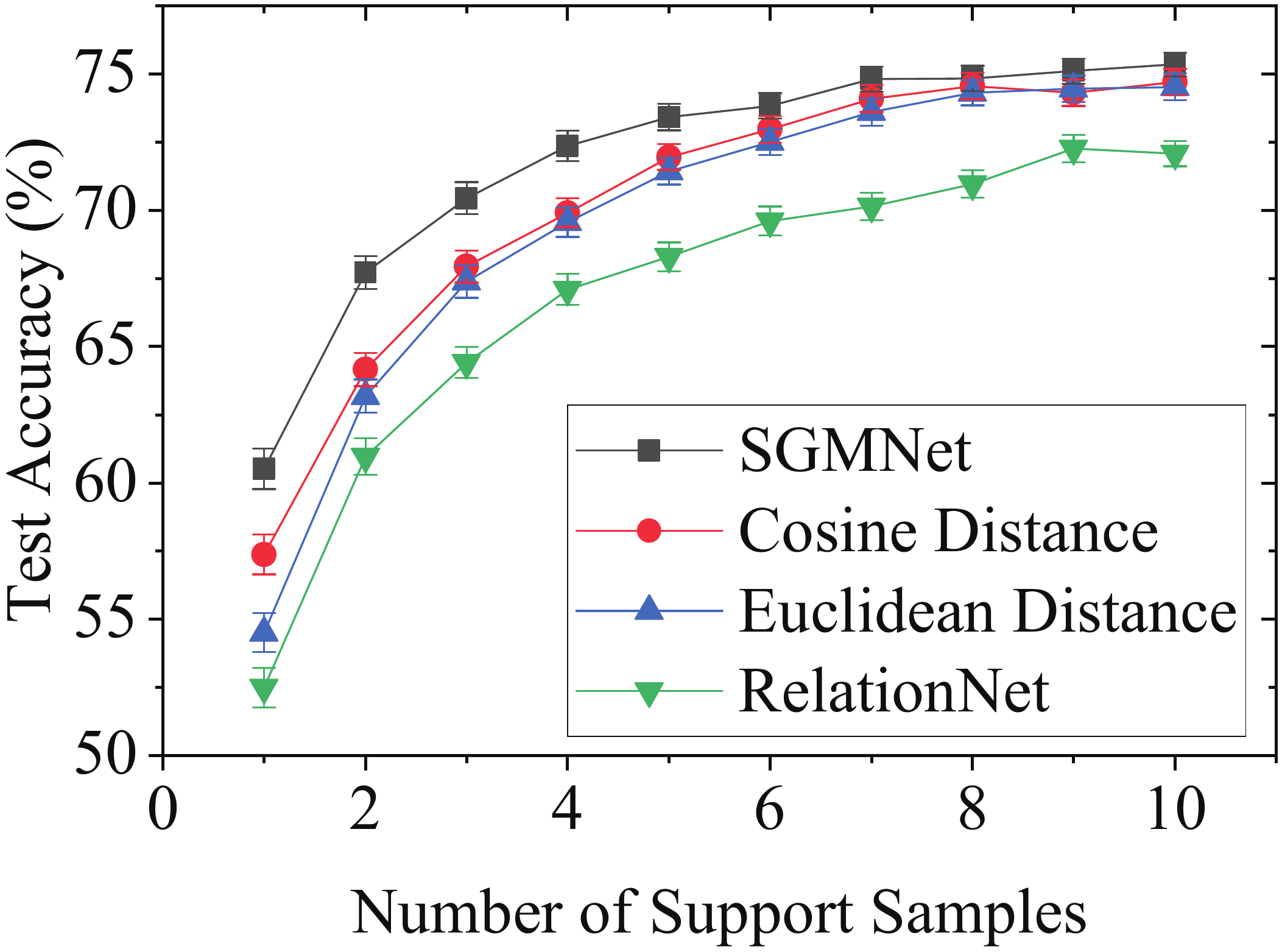}}
	\subfigure[WHU-RS19]{ 
		\label{fig4b} 
		\includegraphics[width=0.48\columnwidth]{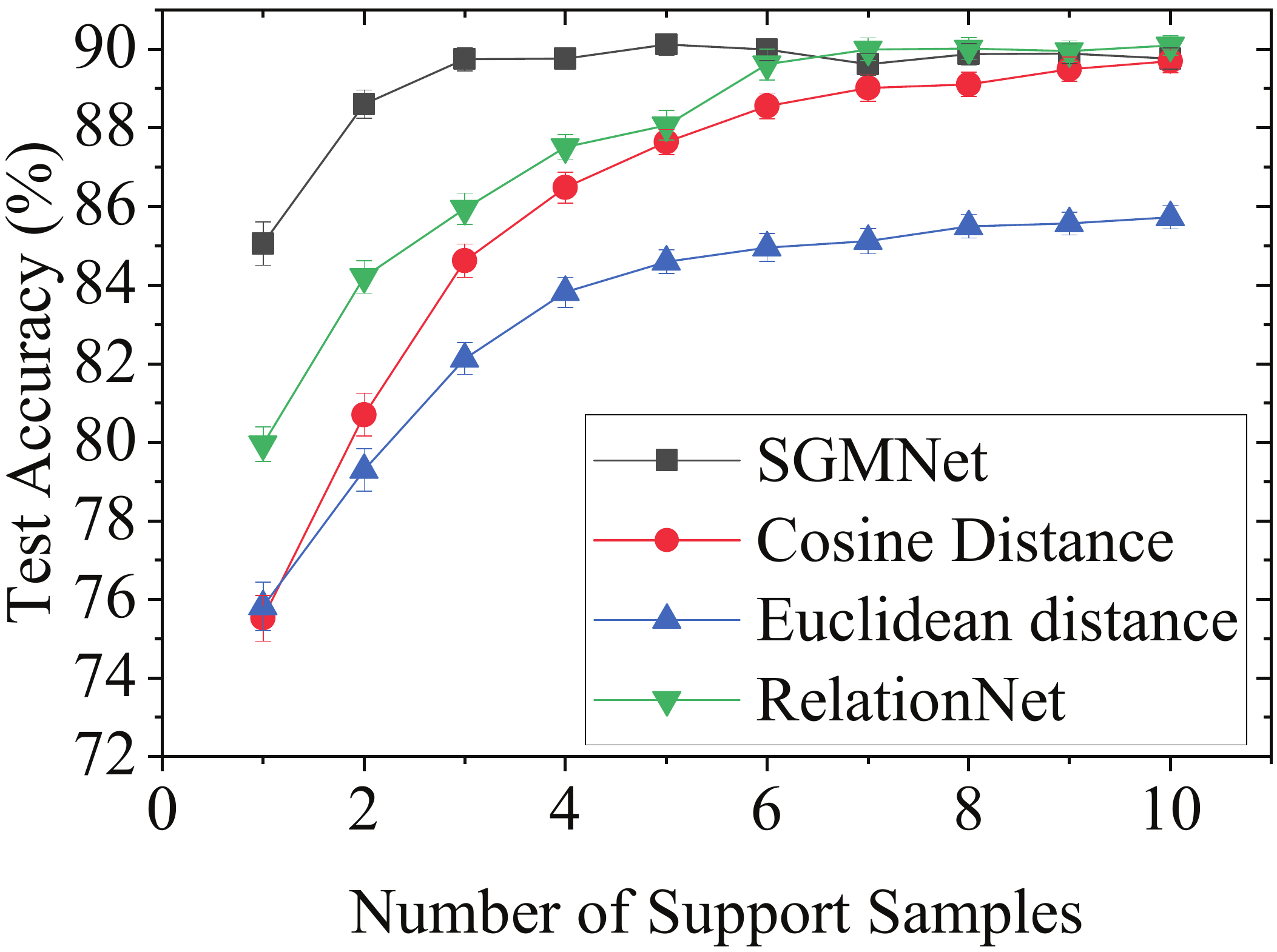}}
	\subfigure[AID]{ 
		\label{fig4c} 
		\includegraphics[width=0.48\columnwidth]{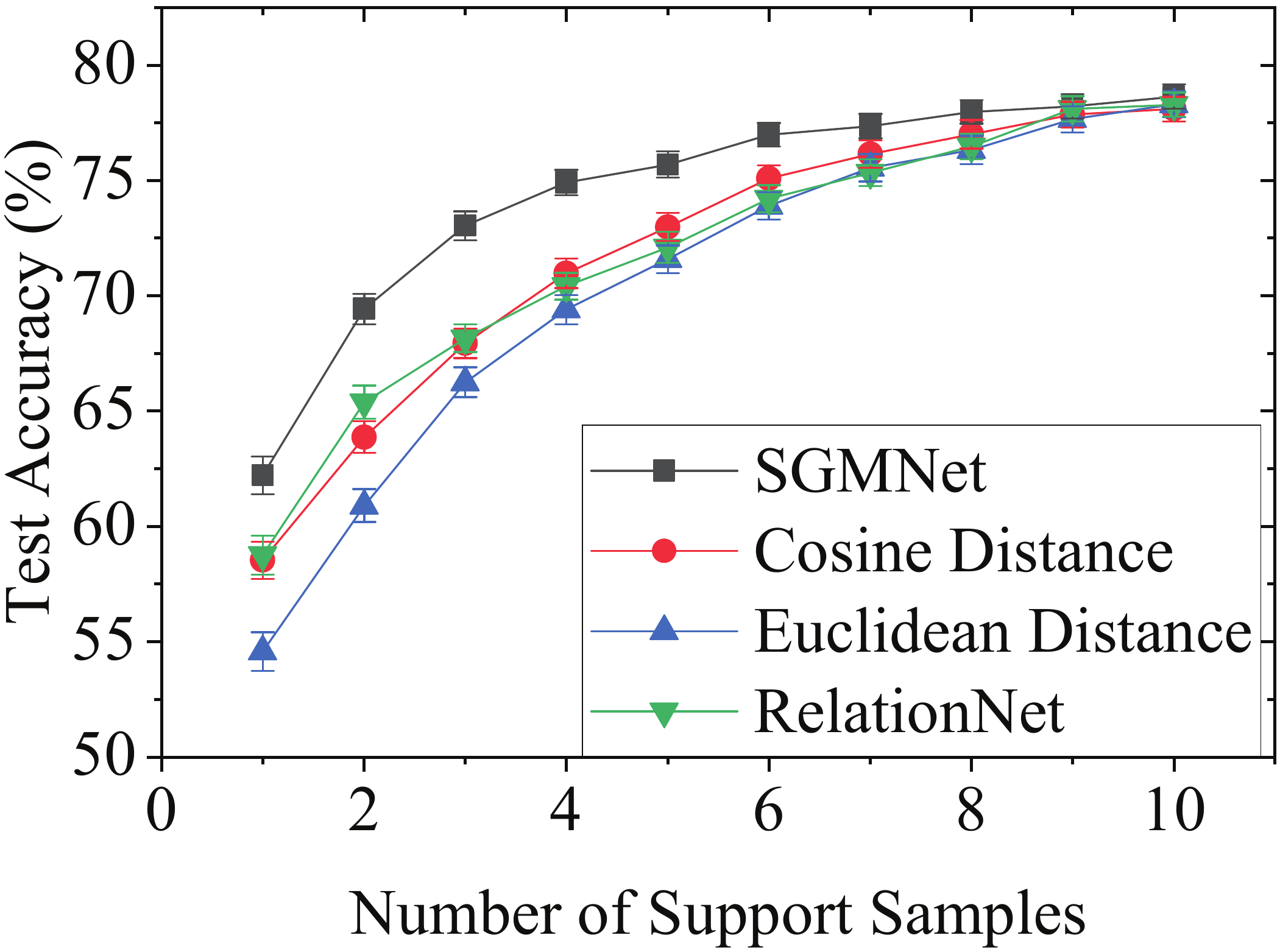}}
	\subfigure[NWPU-RESISC45]{ 
		\label{fig4d} 
		\includegraphics[width=0.48\columnwidth]{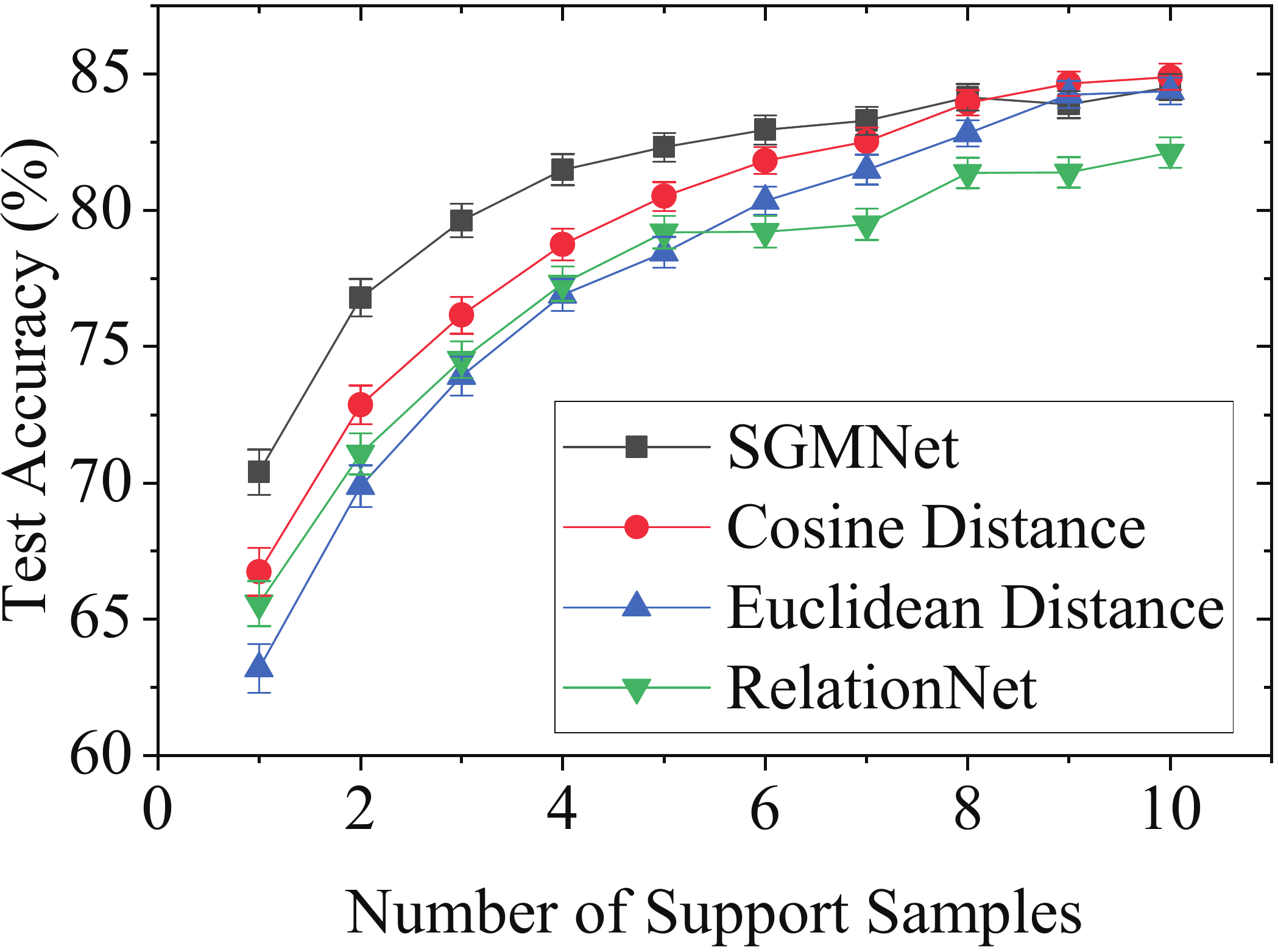}}
	\caption{Experiment results of 5-way FSRSSC task setting with different number of support samples on the four real-world datasets.}
	\label{fig4}
\end{figure*}

\begin{figure*}
	\centering
	\subfigure[UCMerced LandUse]{ 
		\label{fig5a} 
		\includegraphics[width=0.48\columnwidth]{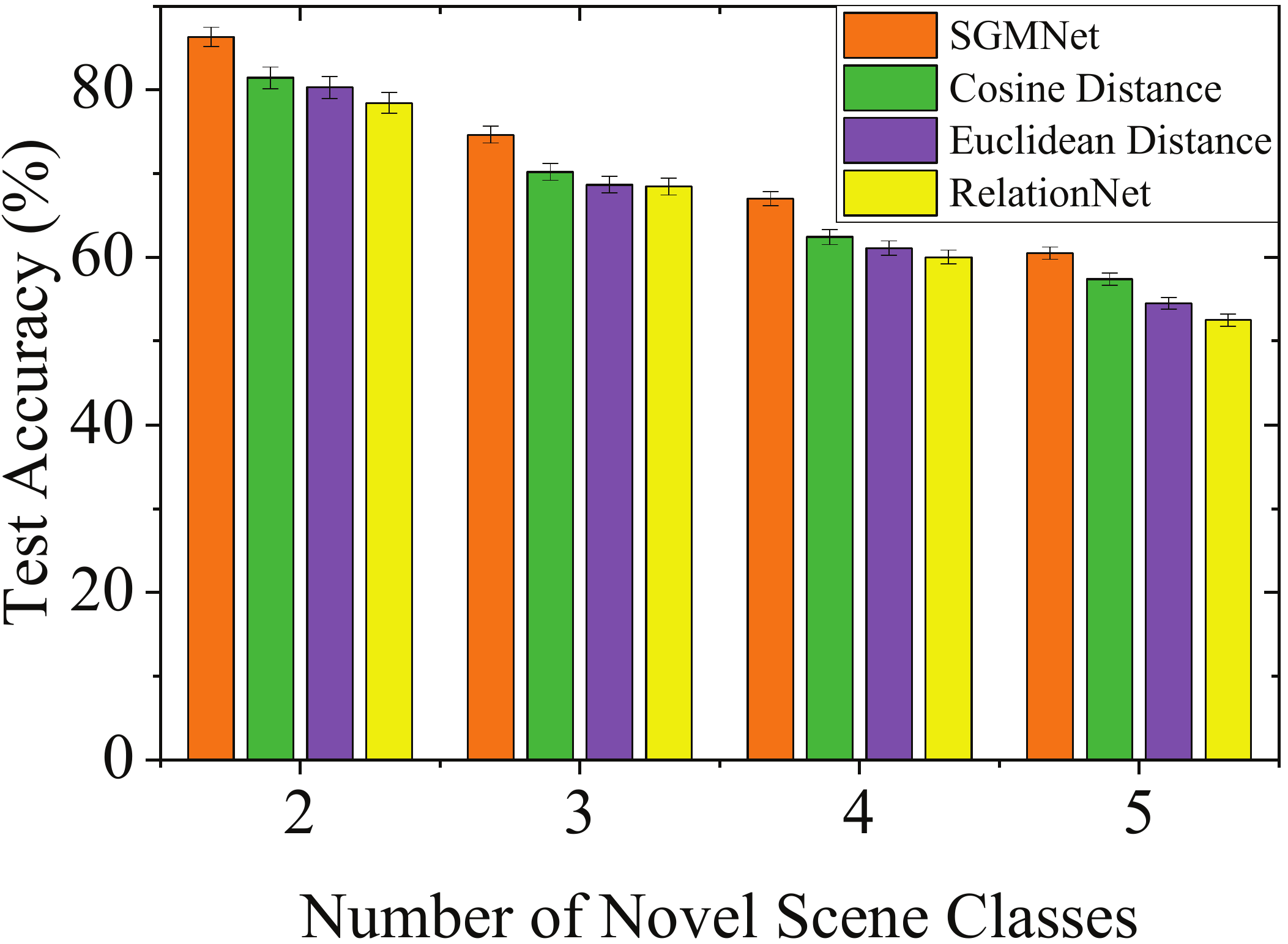}}
	\subfigure[WHU-RS19]{ 
		\label{fig5b} 
		\includegraphics[width=0.488\columnwidth]{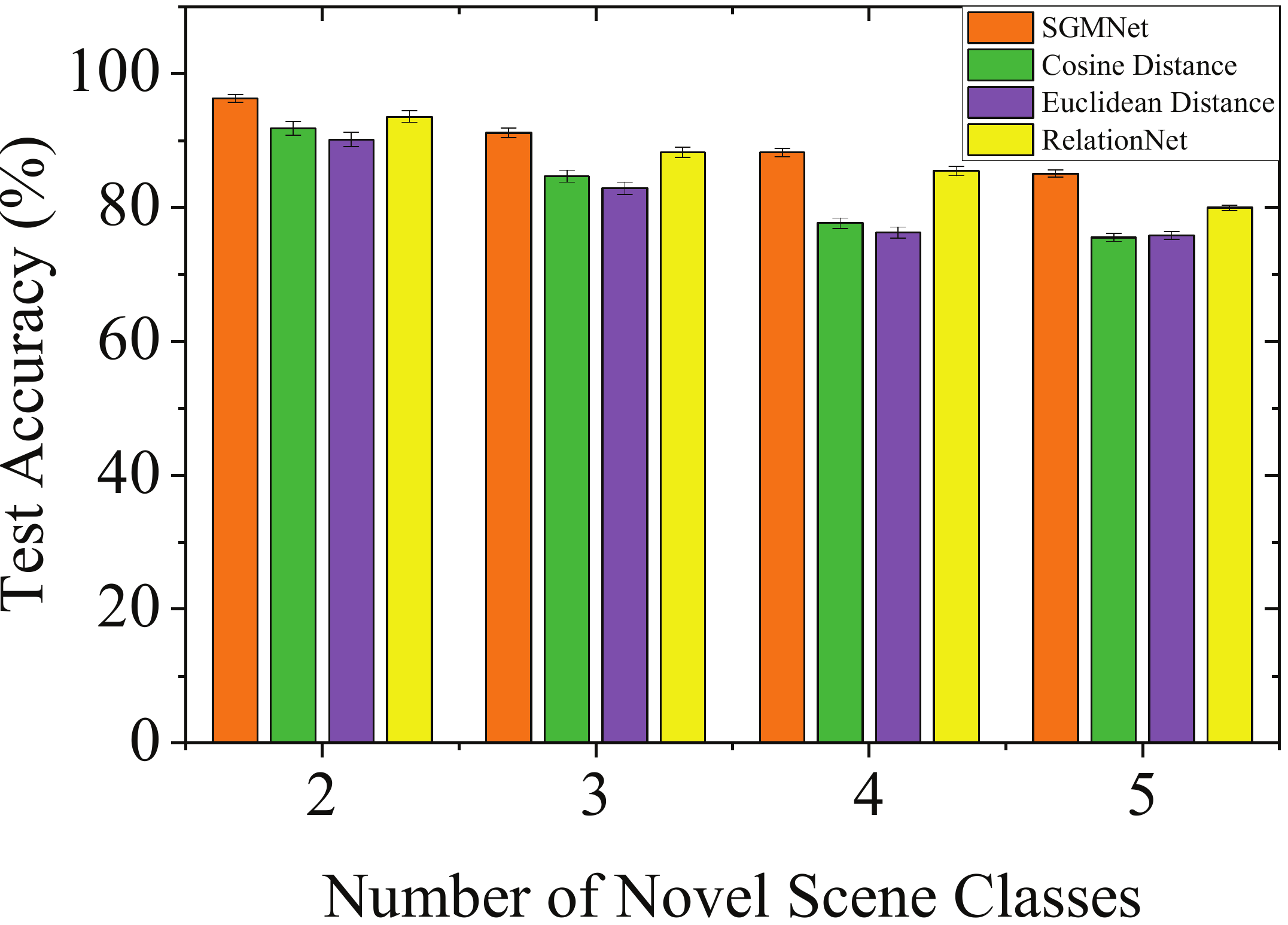}}
	\subfigure[AID]{ 
		\label{fig5c} 
		\includegraphics[width=0.48\columnwidth]{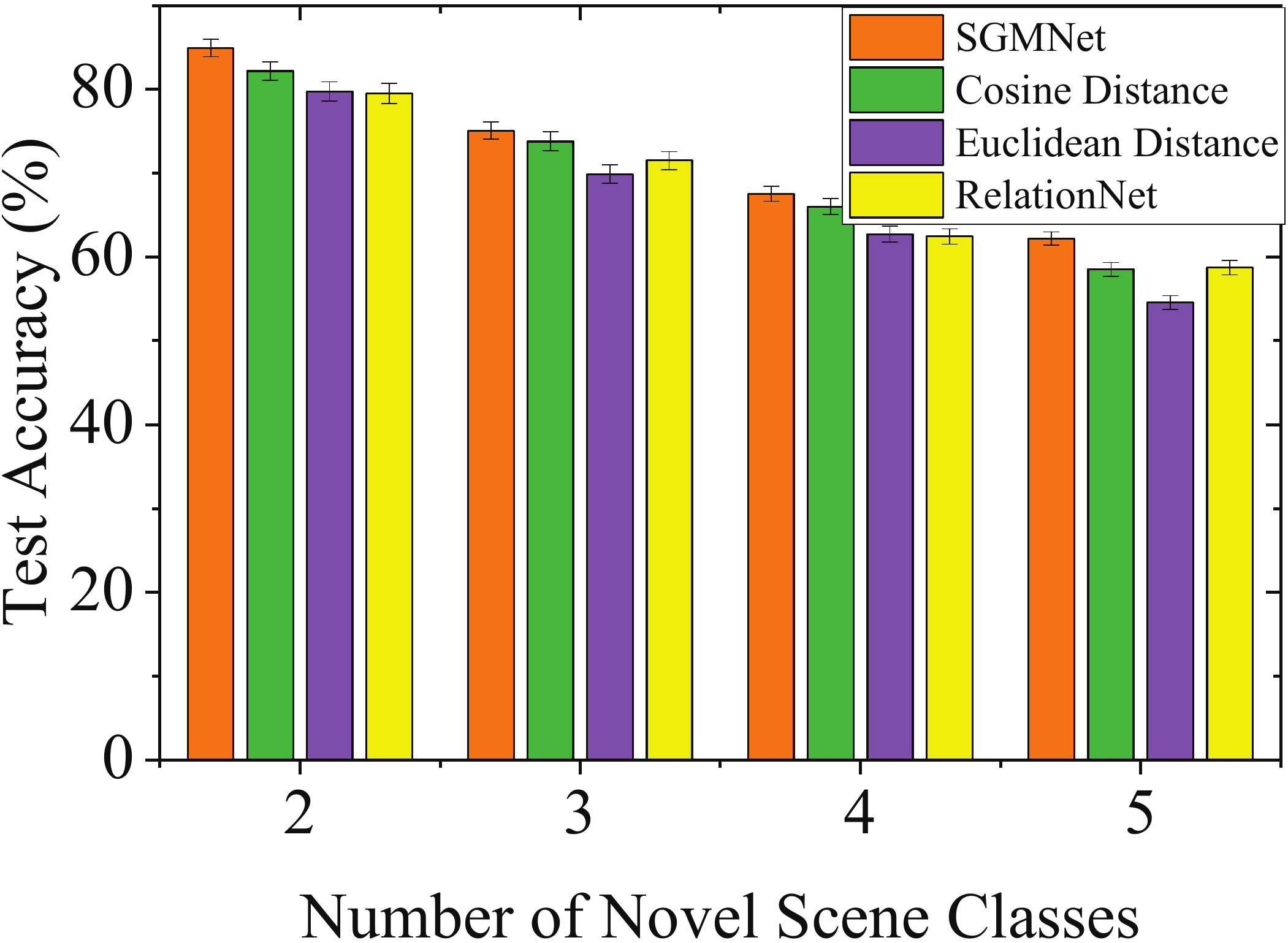}}
	\subfigure[NWPU-RESISC45]{ 
		\label{fig5d} 
		\includegraphics[width=0.48\columnwidth]{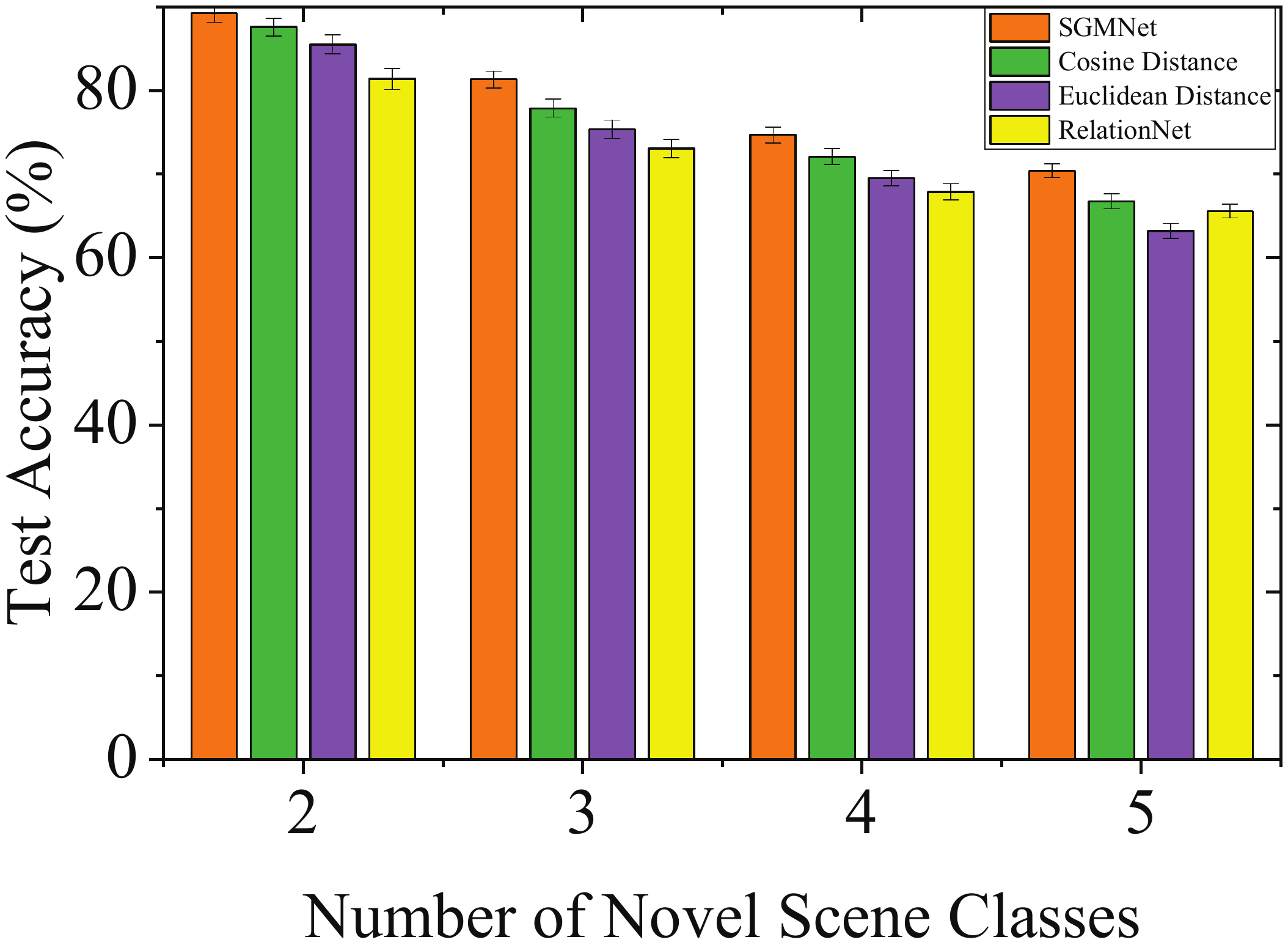}}
	\caption{Experiment results of 1-shot FSRSSC task setting with different number of novel scene classes on the four real-world datasets.}
	\label{fig5}
\end{figure*}

\subsubsection{The Results on Large-Scale Dataset}
Table~\ref{table3} shows the experimental results of the baseline methods and the proposed SGMNet method on AID and NWPU-RESISC45 datasets. We can also observe that our method achieves superior performance with an improvement of 1\% $\sim$ 5\% and 1\% $\sim$ 6\% in 5-way 1-shot and 5-way 5-shot tasks, respectively. This further verifies the effectiveness of our method. Specifically, (\romannumeral1) our method also outperforms these optimization-based methods by around 4\% $\sim$ 10\%. This further shows the superiority of the designed scene graph matching strategy. (\romannumeral2) Compared with the metric-based FSNIC methods (e.g., ProtoNet, MatchingNet, DeepEMD, and RelationNet) and the state-of-the-art FSRSSC methods (e.g., RS-MetaNet and DLA-MatchNet), our method also achieves 3\% $\sim$ 9\% higher accuracy. This further verifies the universality and effectiveness of our SGMNet in the large-scale remote sensing scene dataset, which exhibits richer remote sensing scene classes.

\subsection{Ablation Study}
In this subsection, we conduct the experiments of ablation study to analyze the effectiveness of different-scale backbone, scene graph matching network, two key components (i.e., graph propagation layer and graph interaction layer) of GMM, different support set sizes, different number of novel classes. The experiments are summarized as answering five questions. 

\subsubsection{How does the backbone affect classification performance}
In Table~\ref{table4} and \ref{table5}, we conduct an ablation study to analyze the impacts of different-scale backbone (i.e., Conv-5 and ResNet-12) on the small-scale datasets (UCMercedLandUse and WHU-RS19) and large-scale datasets (AID and NWPU-RESISC45), respectively. It can be found that the classification performance is further improved when a deeper backbone (ResNet-12) is employed in our framework, around 2\% $\sim$ 4\%. This is reasonable because these deep backbones have more learnable parameters and have a higher performance of feature representation for remote sensing scene images. 

\subsubsection{Is the scene graph matching network effective}
In Table~\ref{table6} and \ref{table7}, we evaluate the classification performance of different metric methods on the pre-trained feature extractor, including non-parametric metric (i.e., Euclidean distance and Cosine distance) and learnable metric (i.e., RelationNet). Note that our method employs a scene graph matching network to evaluate the similarity score in the scene graph level. Thus, it can be regarded as a learnable metric method. From Table~\ref{table6} and \ref{table7}, we observe that: (\romannumeral1) compared with these non-parametric metric methods, our method achieves 1\% $\sim$ 3\% higher classification accuracy; (\romannumeral2) our method also beats the learnable metric method (i.e., RelationNet), with an improvement of 5\% $\sim$ 8\%. Different from these metric methods, our method effectively exploits the unique advantages of remote sensing scene images, i.e., object co-occurrence and object spatial correlation, by regarding each remote sensing scene image as a scene graph. These experimental results demonstrate the effectiveness of our method. 

\subsubsection{Is the two key components (graph propagation layer and graph interaction layer) effective in GMM}
In Table~\ref{table8} and \ref{table9}, we evaluate the effect of the two key components by removing them in the proposed SGMNet, respectively. Specifically, (\romannumeral1) we remove the graph propagation layer on GMM and revise the input dimensions of the graph update layer to 512; (\romannumeral2) we remove the graph interaction layer on GMM and revise the input dimensions of the graph update layer to 512. From the results of setting (\romannumeral1) and (\romannumeral2) of  Table~\ref{table8} and \ref{table9}, we observe that the classification performance of the proposed SGMNet decreases by 1\% $\sim$ 2\% when removing the two components, respectively. These experimental results imply that employing the graph propagation layer and the graph interaction layer is helpful for our method.

\begin{figure*}
	\centering
	\includegraphics[width=2.0\columnwidth]{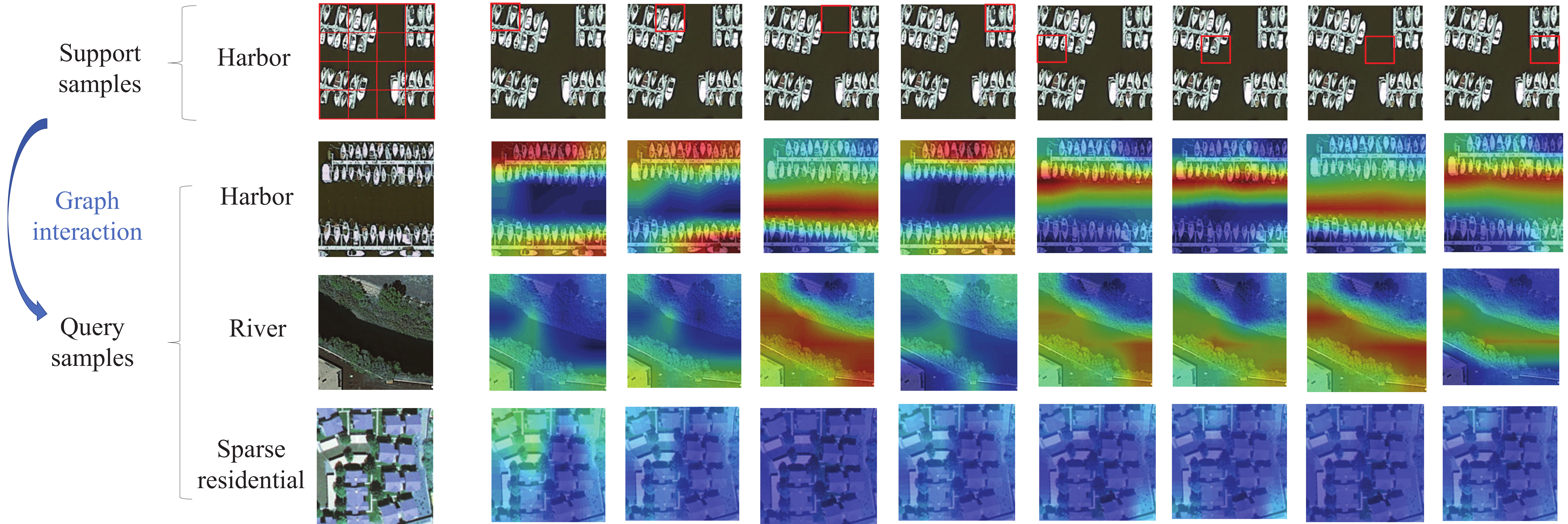}
	\caption{Visualization of graph matching mechanism for SGMNet. The first column represents the original images. The subsequent eight columns show the matching results of each local feature (i.e., each node of scene graph) represented by a red square box in the supporting image with the query images. Note that the matching results are shown by using a bilinear interpolation manner, i.e., upsampling the matching weights of graph interaction layer to the size of query images. We can see that the matching weights can align the same local features (i.e., potential co-occurrence objects like ships and water) well when the support image and the query image match or partially match. Meanwhile, it tends to produce relatively low matching weights when they don’t match. }
	\label{fig7}
\end{figure*} 

\subsubsection{How does the number of labeled samples affect classification performance}
We conduct a statistical experiment on UCMerced LandUse, WHU-RS19, AID, NWPU-RESISC45 datasets by varying the number of labeled samples per class (i.e., $K$) from $K=1$ to $K=10$. The experimental results are shown in Fig.~\ref{fig4}. From Fig.~\ref{fig4}, we find that (\romannumeral1) the classification performance of our method gradually increases with the number increase of labeled samples. This is reasonable because more discriminated features can be obtained from these labeled samples; (\romannumeral2) our method outperforms the existing methods, especially when only few labeled samples are available (e.g., $K=1 \sim 5$); and (\romannumeral3) our method performs slightly worse than the existing methods on WHU and NWPU-RESISC45 datasets when $K=8 \sim 10$. These results implies that exploring object co-occurrence and object spatial correction is more beneficial for FSRSSC, when very few labeled remote sensing images are available (e.g., $K=1 \sim 8$). This is because these unique characteristic can provide more refined description for encoding each scene class, thereby effectively alleviates the data scarcity issue of FSRSSC.

\subsubsection{Is our method effective on different numbers of novel scene classes}
In Fig.~\ref{fig5}, we report the classification performance of our method on 1-shot tasks by varying the number of novel scene classes from $N=2$ to $N=5$. It can be found that (\romannumeral1) the classification performance of our method decreases as the class number $N$ increases; (\romannumeral2) our method outperforms the existing method on all few-shot task settings by around 1\% $\sim$ 10\%. This means that our method is robust and effective for various $N$-way classification task settings.

\subsection{Visualization Analysis}
In this section, we show how the graph matching mechanism works by visualizing the matching weights of the graph interaction layer. The visualization result on the test scene classes of ``harbor'',  ``river'', and ``sparse residential'' from UCMerced LandUse dataset is showned in Fig.~\ref{fig7}. Here, the proposed SGMNet is trained by following 5-way 1-shot task setting. Specifically, in the visualization experiment, we randomly select a remote sensing image from the scene class of ``harbor'' as the support sample, and three remote sensing images from the scene class of ``harbor'', ``river'', and ``sparse residential'', respectively, as the query samples. It is worth noting that the harbor image has two objects of co-occurrence, i.e., ships and water. The river image contains three objects of co-occurrence (i.e., grass, trees and water). Here, the object of ``water'' is overlaped with the co-occurrence objects of the harbor image. In the sparse residential image, there is five objects of co-occurrence, i.e., grass, trees, buildings, cars, and pavement, and we note that none of these five objects appear in the harbor image. Then, we show the matching weights of the graph interaction layer (i.e., the details of matching between each potential object of the support sample and all potential objects of the query samples) in Fig.~\ref{fig7}. Note that each node of scene graph is related to the local feature of the image, which would denote a potential object. From Fig.~\ref{fig7}, we find that the matching weights can align the same objects (i.e., co-occurrence objects such as ships and water) well when the two remote sensing images match or partially match, while tend to produce relatively low matching weights when they don’t match. This visualization results indicate that the proposed graph matching mechanism is helpful for FSRSSC, which can effectively align and highlight the same objects appearing in two remote sensing images and provide more refined matching detail information for performing the similarity evaluation between two remote sensing scene images. 

\section{Conclusion}
\label{sec_conclusion}
In this paper, we identify two unique characteristics of remote sensing images for few-shot remote sensing scene classification (FSRSSC), i.e., object co-occurrence and object spatial correction. Their advantage is that they can provide more refined descriptions for remote sensing scene classes, thereby  alleviating the data scarcity issue of FSRSSC. To fully leveraging the advantage, we propose a novel scene graph matching-based meta-learning framework for FSRSSC. In particular, we regard each remote sensing image or each scene class as a scene graph, where the nodes reflect these co-occurrence objects and the edges capture the spatial correlations between these co-occurrence objects. Then, a novel scene graph matching network is carefully designed to compute the similarity between each test remote sensing image and each scene class at the scene graph level. Finally, based on these similarity scores, we perform the novel scene class prediction through a nearest neighbor classifier. Experiments on four remote sensing datasets show that our model obtains significantly superior performance over state-of-the-art methods. We also conduct extensive ablation studies and visualization analysis, which further show the superiority of our method, especially when only very few labeled samples are available.


%



\section*{Acknowledgment}
This work was supported by the Shenzhen Science and Technology Program under Grant No. JCYJ201805071838- 23045 and Grant No. JCYJ20200109113014456.

\ifCLASSOPTIONcaptionsoff
  \newpage
\fi




\bibliographystyle{IEEEtran}
\bibliography{egbib}

\begin{IEEEbiographynophoto}{Baoquan Zhang}
is currently pursuing the Ph.D. degree with the School of Computer Science and Technology, Harbin Institute of Technology, Shenzhen, China. He received the B.S. degree from the Harbin Institute of Technology, Weihai, China, in 2015, and the M.S. degree from the Harbin Institute of Technology, China, in 2017. His current research interests include meta learning, few-shot learning, and machine learning.
\end{IEEEbiographynophoto}

\begin{IEEEbiographynophoto}{Shanshan Feng}
	is currently an Associate Professor with the School of Computer Science and Technology, Harbin Institute of Technology, Shenzhen, China. He received the Ph.D. degree in Computer Science from Nanyang Technological
	University, Singapre, in 2017. His research interests include sequential data mining and social network analysis.
\end{IEEEbiographynophoto}


\begin{IEEEbiographynophoto}{Xutao Li}
	is currently an Professor with the School of Computer Science and Technology, Harbin Institute of Technology, Shenzhen, China. He received the Ph.D. and Master degrees in Computer Science from Harbin Institute of Technology in 2013 and 2009, and the Bachelor from Lanzhou University of Technology in 2007. His research interests include data mining, machine learning, graph mining, and social network analysis, especially tensor-based learning, and mining algorithms.
\end{IEEEbiographynophoto}


\begin{IEEEbiographynophoto}{Yunming Ye}
	is currently a Professor with the School of Computer Science and Technology, Harbin Institute of Technology, Shenzhen, China. He received the PhD degree in Computer Science from Shanghai Jiao Tong University, Shanghai, China, in 2004. His research interests include data mining, text mining, and ensemble learning algorithms.
\end{IEEEbiographynophoto}

\begin{IEEEbiographynophoto}{Rui Ye}
    is currently pursuing the Ph.D. degree with the Department of Computer Science and Technology, Harbin Institute of Technology, Shenzhen, China. He received the B.S. degree from the University of Electronic Science and Technology of China, Chengdu, China, in 2016, and the M.S. degree from the University of Electronic Science and Technology of China, Chengdu, in 2019. His current research interests include spatiotemporal data mining, time series forecasting, and machine learning.
\end{IEEEbiographynophoto}

\appendices
\section{Details of Datasets}
In the section of experiments, we evaluate some baseline methods and our SGMNet method on four real-world remote sensing datasets, i.e., UCMercedLandUse, NWPU-RESISC45, WHU-RS19, AID, and MLRSNet. Here, for the UCMercedLandUse, NWPU-RESISC45, and WHU-RS19 datasets, we split them into three scene class sets, i.e., meta-training, meta-validation, and meta-test. The split strategy is following the split strategies proposed by \cite{2020DLAmatchnet}. Please refer to their origin paper for more details. It is worth noting that for the AID data set, we use a random manner to split it into 16 scene classes for meta-training, 7 scene classes for meta-validation, and 7 scene classes for meta-test, respectively. Specifically, the meta-training class set contains 16 scene classes, i.e., playground, stadium, resort, mountain, center, port, bridge, storage tanks, park, square, farmland, desert, commercial, railway station, beach, and parking; the meta-validation class set contains 7 scene classes, i.e., church, meadow, baseball field, school, river, bare land, and pond; and the meta-test class set contains 7 scene classes, i.e., viaduct, industrial, dense residential, medium residential, airport, forest, and sparse residential.




\end{document}